%% file: main.tex
\documentclass{article}%
\usepackage{iclr2027_conference,times}

\usepackage[utf8]{inputenc}%
\usepackage[T1]{fontenc}%
\usepackage{hyperref}%
\usepackage{url}%
\usepackage{booktabs}%
\usepackage{amsfonts}%
\usepackage{nicefrac}%
\usepackage{microtype}%
\usepackage{xcolor}%
\usepackage{graphicx}
\usepackage{subfigure}
\usepackage{bm}
\usepackage{dsfont}
\usepackage{amsmath}
\usepackage{amssymb}
\usepackage{mathtools}
\usepackage{amsthm}
\usepackage{xurl}
\usepackage{xspace}

\input{prestuff}

\usepackage{algorithmic}
\usepackage[capitalize,noabbrev]{cleveref}
\usepackage{bbm}

\theoremstyle{plain}
\newtheorem{theorem}{Theorem}[section]
\newtheorem{proposition}[theorem]{Proposition}

\theoremstyle{definition}

\theoremstyle{remark}

\usepackage{enumitem}
\setlist[itemize]{leftmargin=*, itemsep=0pt, topsep=-3pt}
\usepackage{fontawesome5}

\renewcommand{\mcite}[1]{\,\citep{#1}}

\title{Understanding and Mitigating the Fragility of Self-Rewarding Optimization}

\author{Rongyi Zhu \\
Stony Brook University \\
\texttt{rozzhu@cs.stonybrook.edu}
\And
Tanqiu Jiang\\
Stony Brook University \\
\texttt{tanjiang@cs.stonybrook.edu}
\AND
Fenglong Ma \\
Pennsylvania State University \\
\texttt{fenglong@psu.edu}
\And
Ting Wang \\
Stony Brook University  \\
\texttt{twang@cs.stonybrook.edu}
}

\iclrfinalcopy
\begin{document}

\maketitle

\begin{abstract}
Self-rewarding optimization (SRO) promises to improve language models by enabling them to generate their own reward signals. Yet when and how SRO succeeds or fails remains poorly understood. We reveal that SRO is fundamentally fragile: even seemingly benign perturbations to training prompts can corrupt the self-generated preference signal and trigger iterative collapse. We demonstrate this through two prompt-level attacks that serve as controlled stress tests: a Re-Ranking attack that concentrates training on prompts where self-reward diverges from an oracle, and a Preamble attack that learns short prefixes inflating self-reward without any gain in oracle quality. To explain why these attacks succeed, we introduce self-reward fidelity (SRF), a metric quantifying on-policy agreement between self-evaluated and oracle rewards. We show that SRF can serve as a diagnostic signal of self-supervision reliability, and that controlled SRF perturbations induce a phase-transition region: SRO improves when SRF remains sufficiently high, but collapses once SRF drops below a model- and task-dependent threshold. This analysis motivates Projection Decoupling Defense, which suppresses SRO update components along directions to which the reward generator is locally sensitive, improving SRO robustness under adversarial conditions. This work sheds light on the fragility of SRO and charts promising directions toward making it more robust.
\end{abstract}

\section{Introduction}
\label{sec:intro}

Aligning large language models (LLMs) with human intent is typically framed as an optimization problem~\mcite{rafailov2023direct, bai2022training, shao2024deepseekmath}, where the objective is to produce outputs that score highly on quality, correctness, or safety. These alignment pipelines rely on external supervision, such as human preference labels or reward models trained on curated feedback, and optimize the policy via reinforcement learning or preference optimization. While effective, such approaches are expensive to scale and fundamentally limited by the availability, consistency, and coverage of external supervision.

Self-rewarding optimization (SRO)~\mcite{huang2023large,liu2024direct,wu2025meta,zhou2025self,wangcream,wang2025temporal,yuan2024self} offers an appealing alternative. Rather than relying on external reward models, SRO repurposes the LLM itself as both an answer generator and a reward generator, creating a self-contained loop that reduces dependence on human annotations and enables continual self-improvement. Yet empirical results are mixed: the same training recipe can improve performance in some settings while causing sharp degradation or outright collapse in others~\mcite{zhang2025process,yuan2024self,herel2024collapse}. Despite growing interest, the factors governing SRO's effectiveness remain poorly understood.

We begin from this failure mode. If SRO is to serve as a scalable alignment mechanism, it should remain stable under variations in training data, such as reordering, filtering, or adversarial modification of prompts. We show this is not always the case. We introduce two prompt-level adversarial attacks that preserve the SRO training recipe while perturbing only the prompts seen by the self-rewarding loop. The first, a Re-Ranking attack, exploits dataset knowledge to prioritize questions where the reward generator is most unreliable, concentrating training on high-disagreement regions. The second, a Preamble attack, learns a short prompt prefix that systematically inflates self-reward scores without any corresponding oracle improvement, thereby poisoning the preference signal. Across SRO methods~\mcite{self-evolved-reward,wangcream}, LLMs, and tasks, both attacks substantially degrade performance.

These attacks raise a critical question: what property of the SRO pipeline makes such collapses possible? We answer this by analyzing self-reward fidelity (SRF), the on-policy agreement between the model's self-evaluated and oracle rewards. SRF is not an isolated reward-model metric: because the same parameters implement both answer generation and reward generation, answer quality and self-reward fidelity are inherently coupled. By explicitly controlling SRF through targeted reward perturbations while holding the base model and training recipe fixed, we observe a phase-transition region: SRO improves when on-policy reward agreement remains sufficiently high, but degrades as fidelity falls through a model- and task-dependent range.

In response, we propose Projection Decoupling Defense to decouple the effect of answer-policy updates on self-evaluation. We hypothesize that collapse is exacerbated when the same parameters are updated in correlated directions for both answering and judging. We realize this idea using verdict-score gradients on a small probe set to identify directions to which the reward generator is locally sensitive. We then suppress the components of SRO updates along those directions. This targeted intervention breaks harmful feedback loops and substantially improves SRO robustness under adversarial reward conditions, while preserving its benefits in benign settings.

To the best of our knowledge, this is the first work to explore the fragility of SRO through prompt-level adversarial attacks and to chart principled directions for improving its robustness. Our contributions are as follows:
\begin{itemize}
\item We expose the fragility of SRO through Re-Ranking and Preamble attacks that manipulate only the prompt distribution yet reliably drive iterative collapse.
\item We introduce self-reward fidelity (SRF) as a diagnostic signal of self-supervision reliability, and show that controlled SRF perturbations induce a model- and task-dependent phase-transition region governing when SRO improves or collapses.
\item We propose Projection Decoupling Defense, which suppresses SRO update components along directions to which the reward generator is locally sensitive, improving SRO robustness under adversarial conditions.
\end{itemize}

\section{Related Work}
\label{sec:relwk}

{\bf Generative Reward Model.} A generative reward model (GRM) treats evaluation as a generation task rather than outputting scalar rewards. Recent work explores both training and evaluation methods for GRMs. For training, \cite{bai2022constitutional,gunjal2025rubrics,yuan2025curing} replace opaque scalar rewards with principles or checklists that guide reasoning before producing rewards, while \cite{lee2023rlaif,yuan2024self, chen2025rm} use strong LLMs as judges to create preference signals or employ self-rewarding approaches. For evaluation, \cite{shi2024judging,wataoka2024self,chen2024humans, bavaresco2024llms, chehbouni2025neither} identify systematic biases, including position bias, verbosity bias, self-preference, and misinformation oversight, that distort judgments. \cite{raina2024llm,maloyan2025adversarial,tian2025overconfidence,bodhwani2025calibrated} further examine the manipulability of GRMs and the reliability of their confidence estimates.

{\bf Preference Optimization.} Reinforcement learning from human feedback (RLHF)\mcite{NIPS2017_d5e2c0ad, ouyang2022training, bradley1952rank, schulman2017proximal} is a standard approach for aligning language models with human preferences: a reward model is trained from pairwise comparisons and the base model is then optimized against this proxy reward using policy-gradient methods such as Proximal Policy Optimization (PPO). However, proxy rewards can be brittle and costly to curate at scale, motivating direct preference optimization methods\mcite{rafailov2023direct} that bypass explicit reward modeling and optimize policies directly from comparisons. Recent work\mcite{azar2024general, xiao2024cal, lee2025kl} provides unified and calibrated perspectives on these objectives, enabling more stable training. In parallel, AI-feedback approaches\mcite{liu2023statistical,hong2024orpo,ethayarajh2024kto,dong2023raft}, such as Constitutional AI\mcite{bai2022constitutional} and RLAIF\mcite{lee2023rlaif}, reduce reliance on human annotations while preserving the benefits of preference-driven learning.

{\bf Self-Rewarding Optimization.}

Self-rewarding optimization (SRO) replaces human-labeled preferences with model-generated ones and can yield iterative alignment gains\mcite{yuan2024self}. Recent work improves SRO through alternative self-supervision strategies, such as high-confidence self-training and contrastive prompt distillation, and reliability mechanisms including meta-judging, reward-consistency regularization, and temporal decoupling of chosen/rejected samples\mcite{huang2023large,liu2024direct,wu2025meta,zhou2025self,wangcream,wang2025temporal}. However, SRO often fails when self-rewards become exploitable through verbosity biases, reward hacking, or coarse answer-level judging on multi-step reasoning\mcite{zhang2025process,yuan2024self,herel2024collapse}. Co-rewarding\mcite{zhang2026corewarding} attributes such collapse to a self-consistent illusion and stabilizes training with rewards from cross-question agreement or a slowly updated teacher, thereby changing how rewards are constructed. We instead study how policy updates degrade the evaluator through shared parameters and constrain those updates directly, representing a complementary direction.

\section{Exposing SRO Fragility}
\label{sec:attack}

\subsection{Self-Rewarding Optimization}

\begin{wrapfigure}{r}{0.5\textwidth}
\centering
\includegraphics[width=0.47\textwidth]{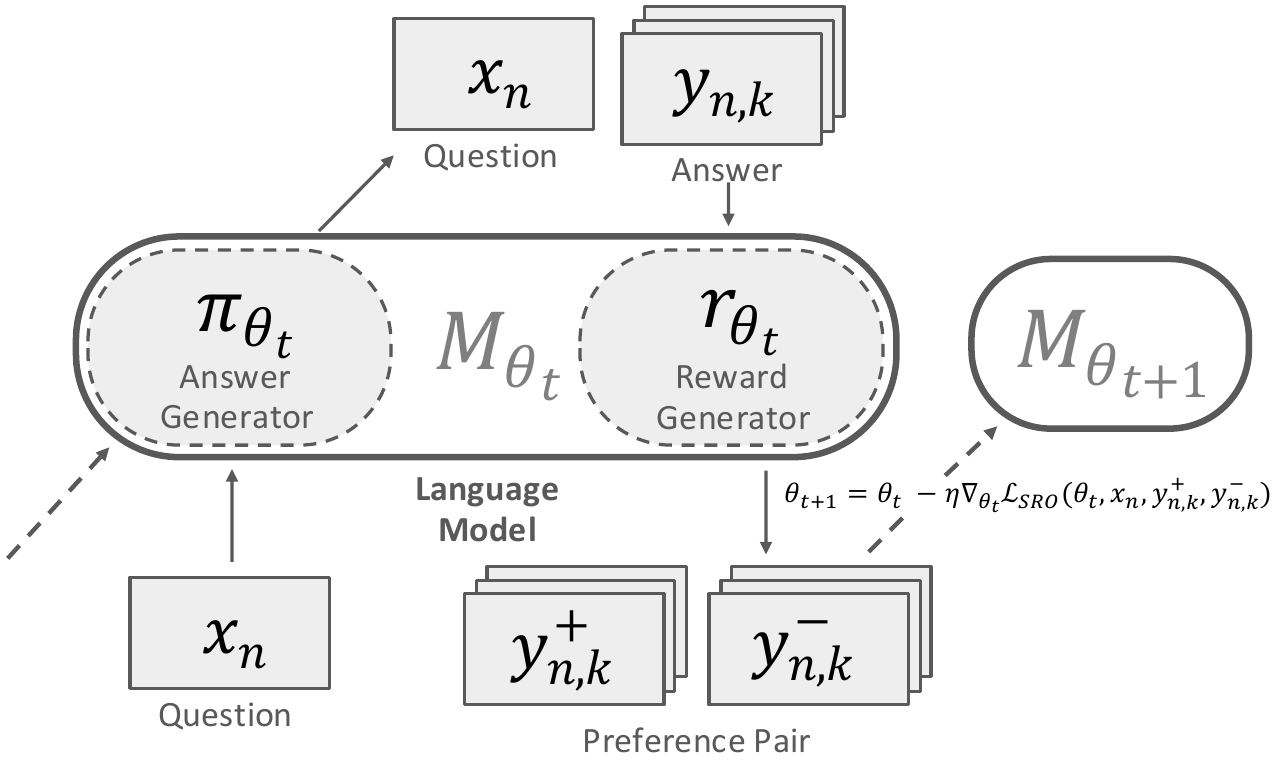}
\caption{\small Self-rewarding optimization (SRO).}
\label{fig:sro}
\vspace{-5pt}
\end{wrapfigure}
As illustrated in Figure\mref{fig:sro}, the LLM $\gM_\theta$ (parameterized by $\theta$) serves both as the answer generator (policy) $\pi_\theta$ and the reward generator $r_\theta$. SRO iteratively uses preference signals produced by $r_\theta$ to update $\pi_\theta$. Specifically, at each iteration $t$, for sampled questions $\{x_n\}_{n=1}^{N}$ drawn from the data distribution $\gD$, the current policy $\pi_{\theta_t}$ generates $K$ candidate
answers $\{y_{n,k}\}_{k=1}^{K}\sim \pi_{\theta_t}(\cdot| x_n)$ for each question; the current reward generator $r_{\theta_t}$ produces self-evaluated rewards $r_{\theta_t}(x_n,y_{n,k})$, which are converted into pairwise preferences ($y^+_{n,k}$, $y^-_{n,k}$); the shared parameters are then updated according to the SRO loss $\mathcal{L}_{SRO}$ with gradient as $g_{SRO}$.

\subsection{Adversarial Attacks on SRO}
To understand the fragility of SRO, we examine how vulnerable SRO is to adversarial manipulations. We consider a threat model wherein the adversaries can only manipulate the prompts (i.e., questions) in the training set. This setting captures practical deployments where externally collected prompts, filtered data pools, or prompt templates can affect which examples enter the self-rewarding loop.
We introduce two novel attack strategies to serve as controlled stress tests of reward fidelit. The proposed attack strategies drive SRO into collapse by manipulating which questions enter the self-rewarding loop. (Detailed attack algorithms are in \msec{app:algo} and additional attack studyis in appendix~\ref{sec:oracle-free-attacks}).

{\bf Re-Ranking Attack.}
We first discuss how to leverage prior knowledge about the prompt dataset to trigger SRO collapse. For a given question $x$, we measure how strongly the reward generator of the target LLM disagrees with the oracle on answers from its current policy $\pi_\theta$. We sample $K$ answers $\{y_i\}_{i=1}^K$ from $\pi_\theta$ and collect rewards from both the oracle $o$ and the reward generator $r_{\theta}$ for each question-answer pair $(x, y_i)$. By comparing these rewards, we define the reward disagreement score of $x$ as:
\begin{equation}
    \kappa = \frac{1}{K}\sum_{i=1}^K\|o(x, y_i) - r_\theta(x, y_i)\|_2,
\label{eq:kappa}
\end{equation}
where higher $\kappa$ indicates greater reward disagreement on answers sampled from the current policy, rather than greater difficulty of $x$ for the answer generator.

By measuring question eward disagreement score, we construct training batches that can trigger SRO collapse. Specifically, given a question set $\mathcal{D}_q=\{x_n\}_{n=1}^{N}$, for each question $x_n \in \gD_q$, we first  compute its reward disagreement score $\kappa_n$ following \meq{eq:kappa}. We then construct a re-ranked dataset by adding Gaussian noise to prevent an overly rigid curriculum. For each $x_n$, we compute a perturbed score $s_n=\kappa_n+\sigma\epsilon_n$ with $\epsilon_n\sim\mathcal{N}(0,1)$ and $\sigma=0.1$. Sorting $\mathcal{D}_q$ in descending order of $s_n$ yields an ordering that concentrates training on questions with large reward-oracle mismatch. Finally, we construct each training batch by sampling $N$ questions from $\mathcal{D}_q$ without replacement according to $s_n$.

{\bf Preamble Attack.}
We further present a Preamble attack that learns short preambles to systematically inflate self-reward scores without improving oracle rewards. To this end, we train an attacker model to generate misleading preambles for each training question. Using DPO, we optimize the attacker model to induce the reward generator to assign inflated scores.%

Specifically, we train a preamble generator $f$ that prepends short text (preamble) to each input prompt to increase the self-reward score while maintaining oracle safety. For each prompt $ x_n$ in the question set $\mathcal{D}_q$, we sample $K$ candidate preambles $p_{n,k}\leftarrow f(\mathcal{I}, x_n)$ (where $\gI$ is the preamble instruction) and generate responses from the fixed policy $\pi_\theta$: a baseline response $y_n\sim \pi(\cdot\mid x_n)$ and preamble-conditioned responses $y^p_{n,k}\sim \pi(\cdot\mid p_{n,k}, x_n)$. We then evaluate their scores using both the reward generator $r_\theta$ and the oracle $o$. The learning signal is the preamble effect on the model's self-reward as
\begin{equation}
A_{n,k} = (r_\theta(p_{n,k}, x_n, y^p_{n,k}) - r_\theta(x_n, y_n)), \quad \text{s.t.} \quad k \in \mathcal{F}_n,
\end{equation}
where $\mathcal{F}_n$ is a feasible set as $\mathcal{F}_n = \{k\in[K]:o(x_n, y^p_{n,k})=o(x_n, y_n)\}$. We update $f$ to maximize the expected advantage under prompts, sampled preambles, and policy rollouts, yielding a trained attacker $f^\star$. We construct the preamble dataset by sampling one preamble per prompt from $f^\star$ and prepending it to the original prompt: $\mathcal{D}_{(p,x)}=\{(p_n,x_n)\}_{n=1}^{N}$. We use $\mathcal{D}_{(p,x)}$ as the input distribution for SRO, disrupting training with questions augmented by preambles designed to exploit the reward model.

\subsection{Experimental Setting}
We empirically evaluate the effectiveness of both attacks, with the following setting.

\textbf{LLMs.} We primarily use four models from two families: {\tt Llama-3.2-3B} and {\tt Llama-3.1-8B} from the {\tt Llama} family, and {\tt Qwen-3-4B} and {\tt Qwen-2.5-7B} from the {\tt Qwen} family.

\textbf{Datasets.} We consider three distinct tasks: general knowledge on MMLU-Pro\mcite{wang2024mmlu}, logic reasoning on ARC\mcite{clark2018think}, and safety alignment on SafeRLHF\mcite{ji2024pku}. From each dataset, we only use questions for SRO. We use {\tt gpt-5} as the oracle evaluator to assess both model-generated answers and rewards.

\textbf{SRO Methods.}
We evaluate two representative SRO methods: \mct{i} Self-Rewarding Language Model (SRLM)\mcite{yuan2024self}, which improves the LLM by generating answers and training on the model's own judgments iteratively, and \mct{ii} CREAM\mcite{wangcream}, which uses cross-iteration ranking agreement to weight DPO and reverse-DPO, stabilizing SRO.

\textbf{Metrics.} We measure answer quality (AQ), the expected oracle reward for model-generated answers:
\begin{equation}
    \text{AQ}(\theta) =
\sE_{x \sim \gD, y \sim \pi_\theta (\cdot| x)} o(x, y),
\label{eq:aq}
\end{equation}
During training, we set temperature as 0.7 and top-$p$ = 0.9 for the answer generator $\pi_\theta$ and temperature as 0.1 and top-$p$ = 0.9 for the reward generator $r_\theta$. We set the learning rate to 1$e$-6 and use the Adam optimizer with a cosine scheduler and warm-up ratio of 0.1. For evaluation, we set temperature as 0.1 and top-$p$ = 0.9 for both $\pi_\theta$ and $r_\theta$.

\subsection{Attack Results}
\label{subsec:results}

\begin{figure}[!t]
\begin{center}
\centerline{\includegraphics[width=\columnwidth]{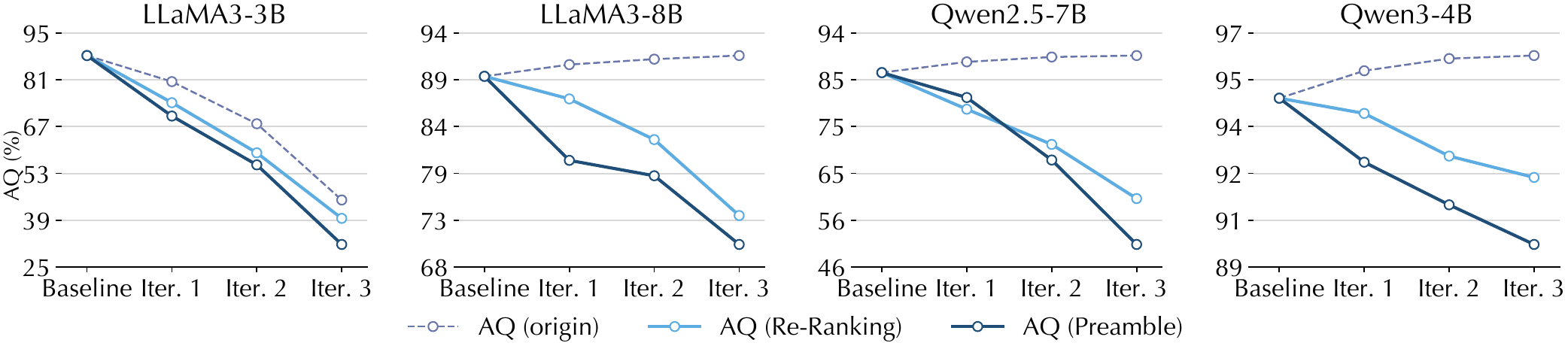}}
\centerline{\includegraphics[width=\columnwidth]{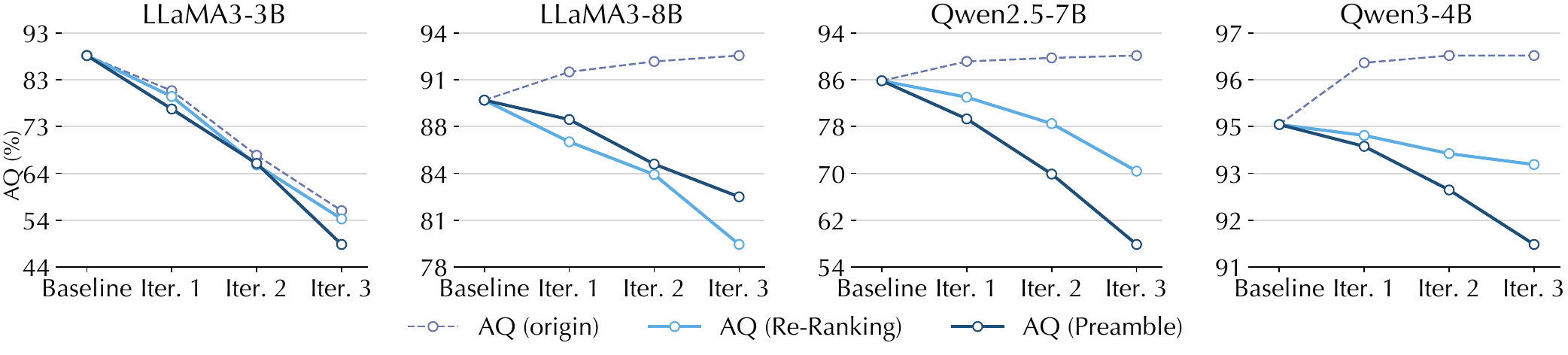}}
\caption{\small Answer quality (AQ) of varied LLMs on the safety alignment task under the Re-Ranking and Preamble attacks (upper: SRLM\mcite{yuan2024self}; lower: CREAM\mcite{wangcream}). Notably, SRO, using either SRLM or CREAM, collapses under prompt-level adversarial attacks.}
\label{fig/exp_results_sro_attack_safty}
\end{center}
\end{figure}

\cref{fig/exp_results_sro_attack_safty} shows that SRO can easily fail under prompt-level adversarial attacks. On safety alignment, clean SRLM is stable for larger models, but this stability disappears once the prompts are perturbed; similar trends on general knowledge and logical reasoning appear in \msec{app:attack}. Re-Ranking and Preamble change only the prompt distribution, yet they drive large AQ drops among models and SRO methods. For SRLM, after three iterations, Llama-3.2-3B’s AQ decreases from 88.6\% to 39.9\% under Re-Ranking and to 32.1\% under Preamble.

\section{Explaining SRO Fragility}
\label{sec:analysis}

\subsection{Self-Rewarding Fidelity}

We first introduce the metric of self-reward fidelity (SRF) as a diagnostic signal of the reliability of self-supervision for SRO. Intuitively, SRF measures the on-policy agreement between rewards from the reward generator $r_\theta$ of $\gM_\theta$ and rewards from the oracle reward generator $o$. Formally,%
\begin{equation}
\label{eq:srq}
\text{SRF}(\theta) =
\sE_{x \sim \gD, y \sim \pi_\theta (\cdot| x)}[ 1 - |o(x, y) - r_\theta(x,y)|],
\end{equation}
where $o(x,y), r_\theta(x,y)\in[0,1]$ denote the normalized rewards from the oracle reward generator and the reward generator of $\gM_\theta$, respectively.
In the binary case, \meq{eq:srq} can be simplified as:
\begin{equation}
\label{eq:srq2}
\text{SRF}(\theta) =
\sE_{x \sim \gD, y \sim \pi_\theta (\cdot| x)}\mathbbm{1}_{[o(x, y) = r_\theta(x,y)]},
\end{equation}
where $\mathbbm{1}_p$ is an indicator function, which returns 1 if predicate $p$ is true and 0 otherwise. Intuitively, SRF measures the on-policy agreement between the reward generator of $\gM_\theta$ and the oracle reward generator $o$. Because SRF is evaluated on samples $y \sim \pi_\theta(\cdot|x)$, it can be seen as an on-policy measure of self-evaluation reliability. It measures whether the model can reliably judge the kinds of answers its current policy produces. Thus, SRF also reflects the compatibility between the answer distribution induced by $\pi_\theta$ and the reward judgments produced by $r_\theta$. SRF and AQ depend on the same model but are not monotone transforms. We use SRF as an oracle-assisted diagnostic signal

\begin{figure}[!t]
\begin{center}
\centerline{\includegraphics[width=\columnwidth]{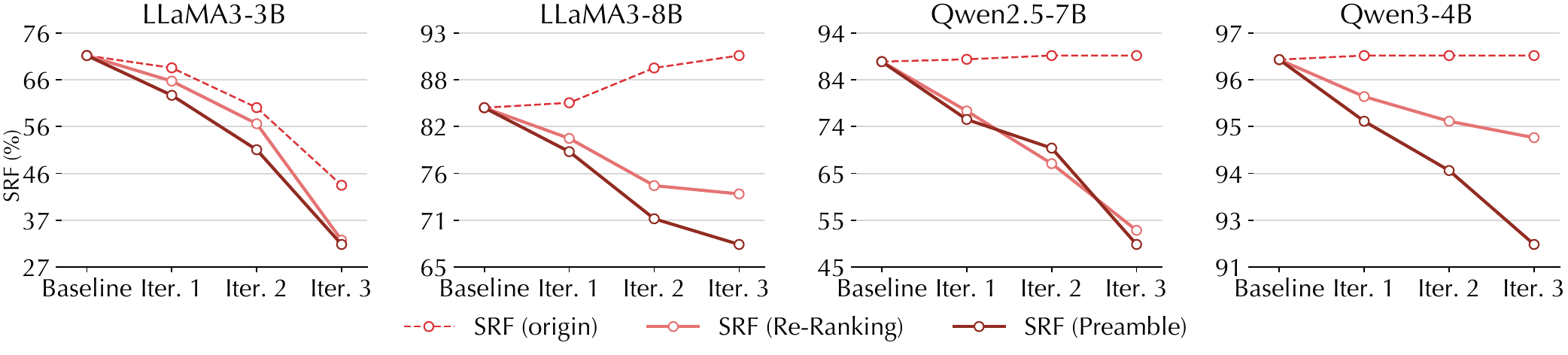}}
\centerline{\includegraphics[width=\columnwidth]{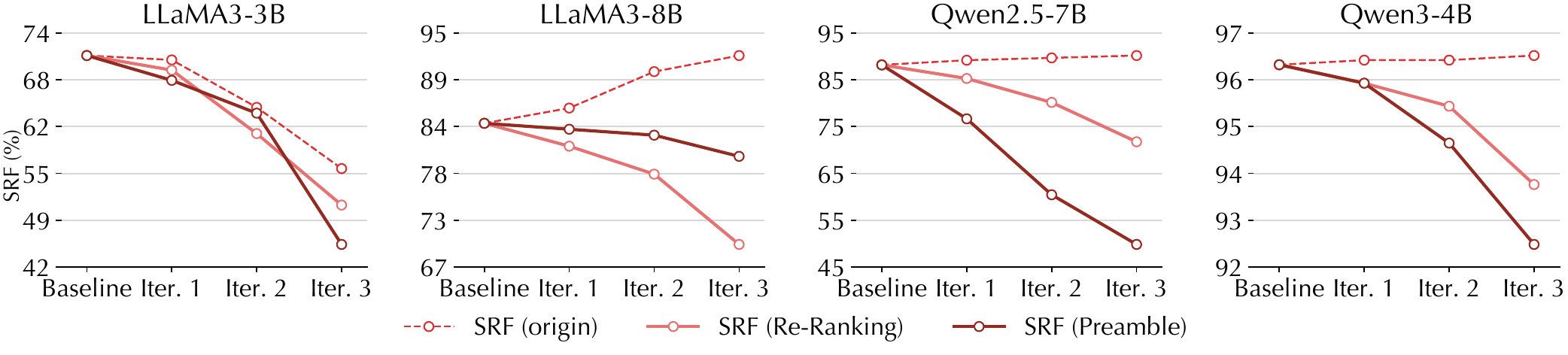}}
\caption{\small SRF broadly tracks the (un)success of SRO. SRF of varied LLMs on the safety alignment task under the Re-Ranking and Preamble attacks (upper: SRLM\mcite{yuan2024self}; lower: CREAM\mcite{wangcream}).}
\label{fig/exp_results_sro_attack_safty_srf}
\end{center}
\end{figure}

\begin{figure}[!t]
\begin{center}
\centerline{\includegraphics[width=\columnwidth]{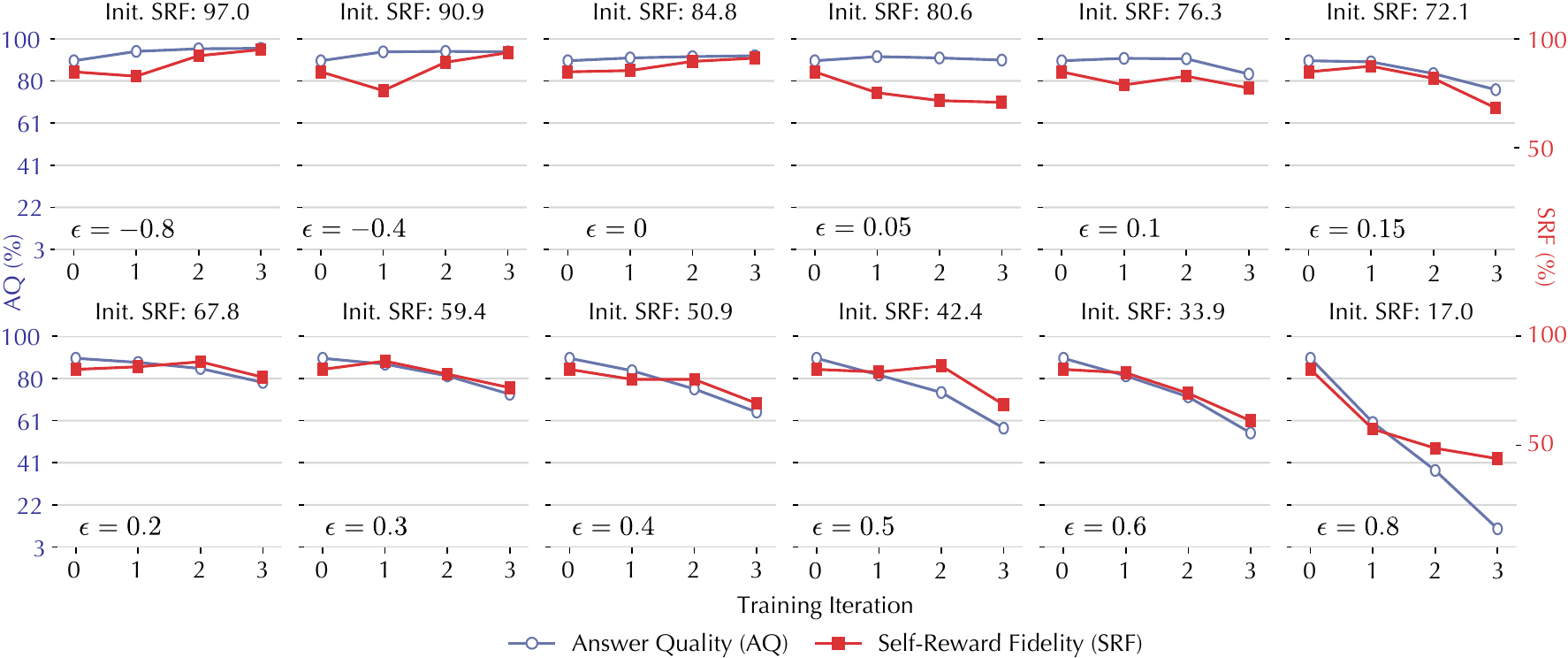}}
\caption{\small Impact of SRF perturbation on SRO dynamics for training {\tt Llama-3.1-8B} on safety alignment: across a phase-transition region, SRO shifts from performance improvement to degradation as initial SRF drops. }
\label{fig/transition_of_noise}
\end{center}
\end{figure}

\begin{figure}[!t]
\begin{center}
\centerline{\includegraphics[width=\columnwidth]{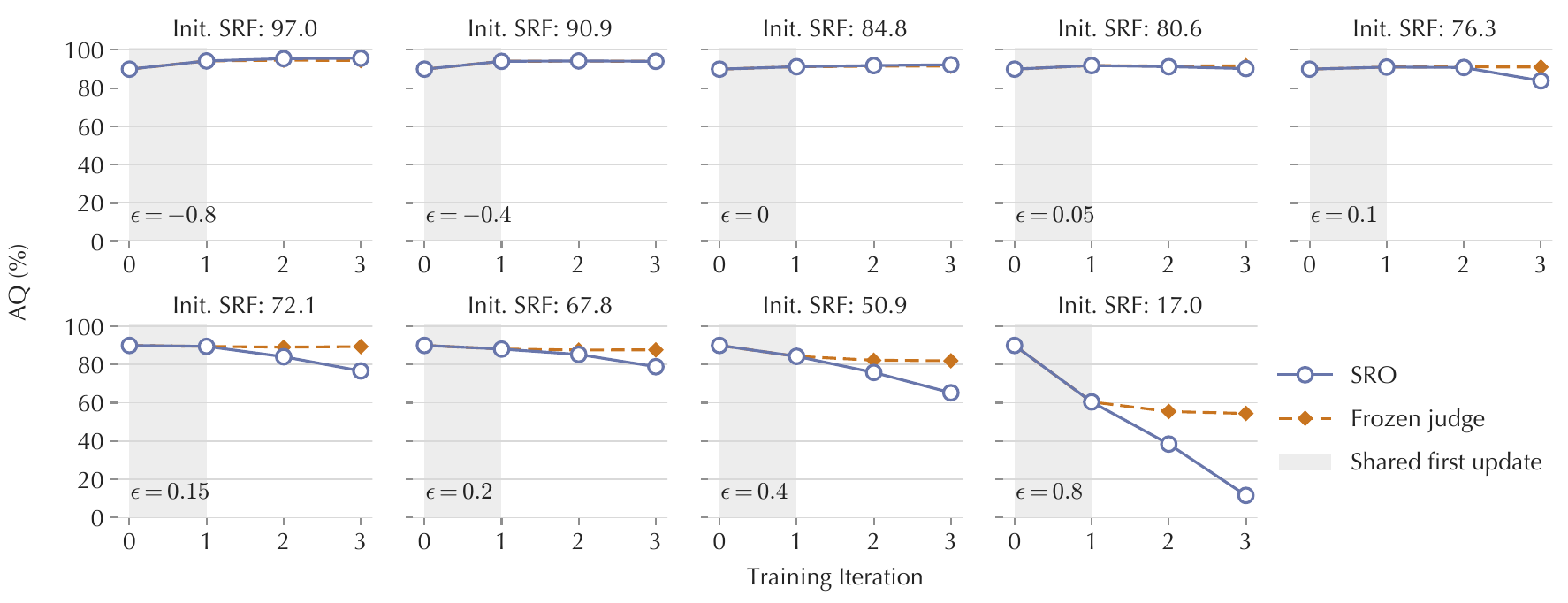}}
\caption{\small {AQ of SRO and a frozen-judge control at the controlled initial SRF levels of \cref{fig/transition_of_noise} ({\tt Llama-3.1-8B}, safety alignment). Frozen judge and SRO  share the first update (shaded), which measures the immediate harm of corrupted labels; the gap that opens afterwards measures its amplification through self-reward feedback.}}
\label{fig:frozen-judge-trajectories-main}
\end{center}
\end{figure}

\subsubsection{SRO Collapse through SRF Lens}

We first show that SRF serves as a useful diagnostic of attack success (\msec{subsec:results}). \cref{fig/exp_results_sro_attack_safty_srf} reports SRF trajectories corresponding to the attack results in \cref{fig/exp_results_sro_attack_safty}. SRF trajectories closely track AQ behavior: clean SRO succeeds when SRF is maintained or improved, while attacked or fragile runs exhibit progressive SRF degradation. This pattern confirms SRF as a diagnostic signal for SRO: high self-reward fidelity permits iterative self-improvement to proceed, whereas declining SRF signals corruption of the self-generated preference signal and portends imminent collapse.

Next, using SRF as a control mechanism, we explore the possible cause of SRO's fragility. To isolate the impact of SRF, we fix the underlying LLM and manipulate SRF by perturbing reward labels at a specified noise ratio $\epsilon$.
To this end, we construct a perturbed training set by flipping a subset of self-reward labels. Specifically, we sample $z\sim \gU(0,1)$ and flip each reward according to the event $z< |\epsilon|$, yielding an expected flip rate $\sP[z < |\epsilon|]= |\epsilon|$. In particular, for $\epsilon >0$, we inject errors by selecting a sample where the self-reward agrees with the oracle ($r_\theta(x,y)=o(x,y)$) and flipping the label to its complement $1-r_\theta(x,y)$. For $\epsilon \le 0$, we instead target a mismatched example ($r_\theta(x,y)\neq o(x,y)$) and flip its label; for binary rewards, this acts as a purification operator that corrects a controllable fraction of incorrect labels. Formally, for $z \sim \gU(0, 1)$, we have:
\begin{equation}
\begin{cases}
\{r_\theta(x,y)\}
& \text{if } z \geq |\epsilon| \\
\{1-r_\theta(x,y) \mid r_\theta(x,y) = o(x,y) \} & \text{if } \epsilon > 0 \wedge z < |\epsilon| \\
\{1- r_\theta(x,y) \mid r_\theta(x,y) \neq o(x,y) \}& \text{if } \epsilon \le 0 \wedge z < |\epsilon|
\end{cases}
\label{eq/noise_eq}
\end{equation}
We measure the impact of SRF perturbation on SRO in \cref{fig/transition_of_noise}, training {\tt Llama-3.1-8B} for safety alignment. We report the initial SRF at each noise level after perturbation. Notably, when $\epsilon <$ 0 (denoising regime), SRO improves both metrics: AQ rises from 89.9\% to 95.6\%, and SRF increases from 84.8\% to 95.0\%. At $\epsilon$ = 0 (no noise), both metrics remain high across iterations, with AQ increasing from 89.9\% to 92.2\% and SRF from 84.8\% to 91.1\%. In contrast, even mild noise ($\epsilon > 0$) degrades performance and halts improvement. At $\epsilon$ = 0.6, AQ decreases to 55.6\% and SRF to 61.2\%. At $\epsilon$ = 0.8, collapse occurs: AQ falls from 89.9\% to 11.6\% and SRF from 84.8\% to 43.8\%.

 By manipulating SRF, we observe a phase-transition region over which SRO shifts from improvement to degradation as initial SRF decreases. In this model- and task-dependent setting, SRO improves AQ when initial SRF is at least 84.8\% ($\epsilon \leq 0$). Across roughly 80.6--72.1\% ($\epsilon$ = 0.05--0.15), the AQ change after three iterations falls from $+0.3$ to $-6.1$ to $-13.3$ points. Below this region, SRO degrades and eventually collapses (AQ 11.6\% at 17.0\% initial SRF). This pattern also holds for {\tt Llama-3.2-3B}, though the region shifts (\msec{sec:pt_llama3b}). Thus, a high SRF alone does not guarantee successful self-improvement. A phase-transition region determined by model and task distribution.

A frozen-judge control (\cref{fig:frozen-judge-trajectories-main}), details in \msec{sec:frozen-judge-srf}, separates the harm of corrupted labels from its amplification by the self-rewarding loop: after a matched first update, AQ falls by a further $12.8$ points by Iteration~3 under SRO versus $0.1$ with a frozen judge at $72.1\%$ initial SRF ($48.8$ versus $6.0$ at $17.0\%$). Thus, evaluator co-evolution substantially amplifies label corruption. Overall, SRF serves as a signal of self-supervision reliability.

\subsection{SRO Collapse through Analytical Lens}
\input{analysis.tex}

\section{Mitigating SRO Fragility}
\label{sec:defense}

\subsection{Answer-Reward Decoupling}

The analysis above suggests that corrupted answer updates can alter the reward generator through shared parameters, reinforcing errors over successive SRO iterations. This motivates a defense that limits the effect of policy updates on reward generation. We propose Projection Decoupling Defense, which attenuates update directions that affect verdict scores under a linear approximation constructed at the start of each SRO round, while retaining directions in its null space.

Specifically, let $\theta\in\mathbb{R}^d$ denote the shared parameters of the answer generator $\pi_\theta$ and reward generator $r_\theta$. We fix a small probe set $\mathcal{C}=\{(x_i,y_i)\}_{i=1}^{N_{\mathcal{C}}}$ of question--answer pairs, with answers sampled from the initial answer generator. For each probe $(x_i,y_i)$, let $\ell_i(\theta)\in\mathbb{R}$ denote its differentiable verdict score: the reward generator's log-probability difference between positive and negative verdicts. We give the full definition in \msec{app:defense_score}.

\begin{figure}[!ht]
\begin{center}
\centerline{\includegraphics[width=\columnwidth]{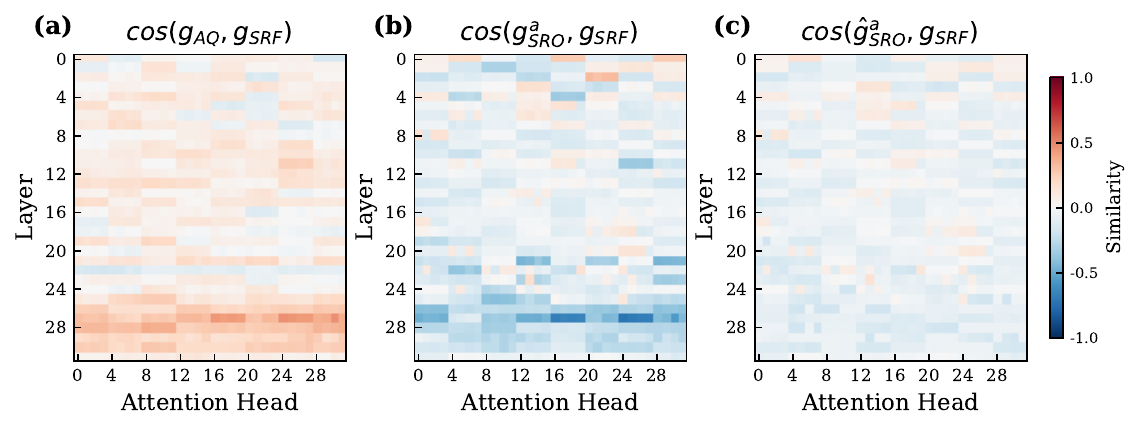}}
\caption{\small Cosine similarity with the self-reward fidelity gradient $g_{\mathrm{SRF}}$ across layers and attention heads: (a) answer-quality gradient $g_{\mathrm{AQ}}$; (b) SRO gradient under the Re-Ranking attack $g^a_{\mathrm{SRO}}$; (c) defended gradient $\hat{g}^a_{\mathrm{SRO}}$. Projection Decoupling Defense reduces negative alignment under attack.}
\label{fig/gradient_simi}
\end{center}
\end{figure}

At the start of each SRO round $t$, we evaluate the verdict-score gradients at $\theta_t$ and define their span:
\begin{equation}
g_i=\left.\nabla_\theta\ell_i(\theta)\right|_{\theta=\theta_t},
\qquad
\mathcal{S}=\operatorname{span}\{g_i\}_{i=1}^{N_{\mathcal{C}}}.
\label{eq:defense-verdict-gradients}
\end{equation}
Let $Q$ have orthonormal columns spanning $\mathcal{S}$. Intuitively, $QQ^\top$ extracts the component of a parameter update that can change the verdict scores under a local linear approximation; the orthogonal component leaves them unchanged to first order. The full basis construction is given in \msec{app:defense_basis}.

During optimization of $\mathcal{L}_{\mathrm{SRO}}$, we recompute the update gradient $g_{\mathrm{SRO}}$. The verdict-score gradients determine which directions to protect, and $g_{\mathrm{SRO}}$ supplies the answer-training update. For exposition, we consider gradient descent with learning rate $\eta$ and defended gradient as
\begin{equation}
\widehat g_{\mathrm{SRO}}=g_{\mathrm{SRO}}-\alpha Q(Q^\top g_{\mathrm{SRO}}),
\qquad
\theta\leftarrow\theta-\eta\widehat g_{\mathrm{SRO}},
\label{eq:fj-update}
\end{equation}
where $\alpha\in[0,1]$ controls protection strength.
\cref{fig/gradient_simi} visualizes gradient alignment across
layers and attention heads. The answer-quality gradient is
generally positively aligned with the SRF surrogate-loss gradient
$g_{\mathrm{SRF}}=\nabla_\theta L_{\mathrm{SRF}}$ (panel~a),
whereas SRO gradients under the Re-Ranking attack exhibit
negative alignment (panel~b). Applying Projection Decoupling
Defense weakens this negative alignment (panel~c), consistent
with reduced local interference with reward generation.

\begin{algorithm}[!ht]\small
\caption{SRO with Projection Decoupling Defense}
\label{alg:fullj}
\KwIn{Initial shared parameters $\theta_1$; training questions $\mathcal{D}_q$; SRO loss $\mathcal{L}_{\mathrm{SRO}}$; candidates per question $K$; SRO rounds $T$; probe count $N_{\mathcal{C}}$; protection strength $\alpha\in[0,1]$; learning rate $\eta$}

Sample $N_{\mathcal{C}}$ questions $x_i$ from $\mathcal{D}_q$ and answers $y_i\sim\pi_{\theta_1}(\cdot\mid x_i)$; set $\mathcal{C}\gets\{(x_i,y_i)\}_{i=1}^{N_{\mathcal{C}}}$\;
\For{$t=1$ \KwTo $T$}{
    Sample $K$ candidate answers per question in $\mathcal{D}_q$ from $\pi_{\theta_t}$\;
    Score the candidates with $r_{\theta_t}$ and construct preference data $\mathcal{D}_t$ following the chosen SRO method\;
    $g_i\gets\left.\nabla_\theta\ell_i(\theta)\right|_{\theta=\theta_t}$ for each fixed probe $(x_i,y_i)\in\mathcal{C}$, using \meq{eq:fj-score}\;
    $J\gets[g_1,\ldots,g_{N_{\mathcal{C}}}]^\top$, $G\gets JJ^\top$\;
    Eigendecompose $G=U\Lambda U^\top$, retaining positive eigenvalues in $\Lambda$ and orthonormal eigenvectors in $U$\;
    $Q\gets J^\top U\Lambda^{-1/2}$\;
    Optimize $\mathcal{L}_{\mathrm{SRO}}$ on $\mathcal{D}_t$ from $\theta_t$ to obtain $\theta_{t+1}$, computing $g_{\mathrm{SRO}}$ at each training step and filtering updates with the fixed $Q$ as in \meq{eq:fj-update}\;
}
\KwOut{Defended shared parameters $\theta_{T+1}$}
\end{algorithm}

\cref{alg:fullj} integrates this update rule into iterative SRO. We first construct the probe set $\mathcal{C}$. In each round $t$, the current answer and reward generators produce preference data $\mathcal{D}_t$, and the verdict-score gradients at $\theta_t$ determine the basis $Q$. We then optimize $\mathcal{L}_{\mathrm{SRO}}$ on $\mathcal{D}_t$, holding $Q$ fixed and filtering each training update as in \meq{eq:fj-update}, to obtain the shared parameters $\theta_{t+1}$ for the next round.

\subsection{Defense Results}

We evaluate Projection Decoupling Defense on safety alignment across the four models in \msec{subsec:results}. All defended runs use $N_{\mathcal{C}}=16$ probes and $\alpha=0.5$. We examine both robustness to Re-Ranking and Preamble attacks and the effect of the defense on attack-free training.

\begin{figure}[!ht]
\begin{center}
\centerline{\includegraphics[width=\columnwidth]{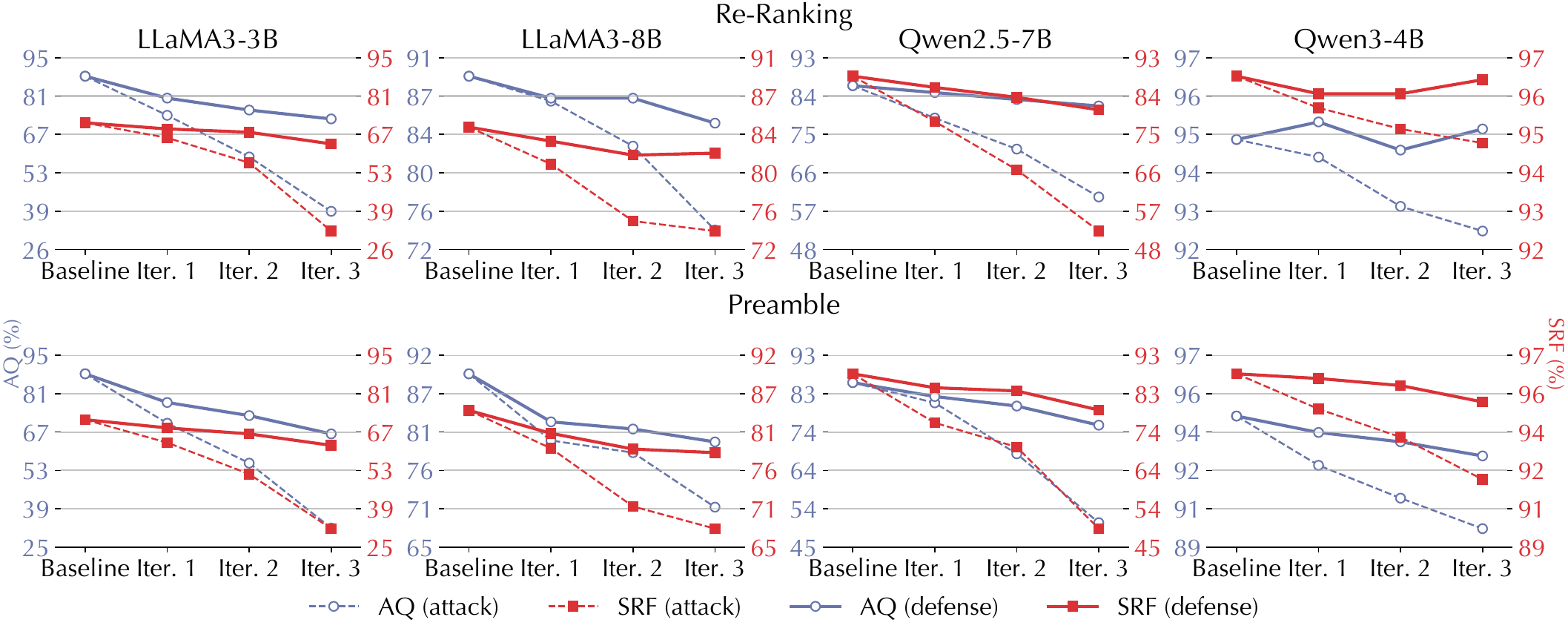}}
\caption{\small Defense against two attacks on SRLM (upper: ReRanking attack; lower: Preamble attack).}
\label{fig/exp_results_defense}
\end{center}
\end{figure}

\paragraph{Robustness under attack.} \cref{fig/exp_results_defense} shows that the defense improves AQ under SRLM by 2.9--34.5 percentage points at iteration~3. Under Preamble, it raises {\tt Llama-3.2-3B}'s AQ from 32.1\% to 66.6\% and SRF from 31.9\% to 62.4\%. These joint gains suggest that limiting the effect of policy updates on reward generation improves SRO robustness under attack. The defense also improves AQ under CREAM by 1.6--24.7 percentage points at iteration~3 (Appendix~\ref{app:defense_cream}, \cref{fig/exp_results_defense_cream}).

\begin{figure}[!ht]
\begin{center}
\centerline{\includegraphics[width=\columnwidth]{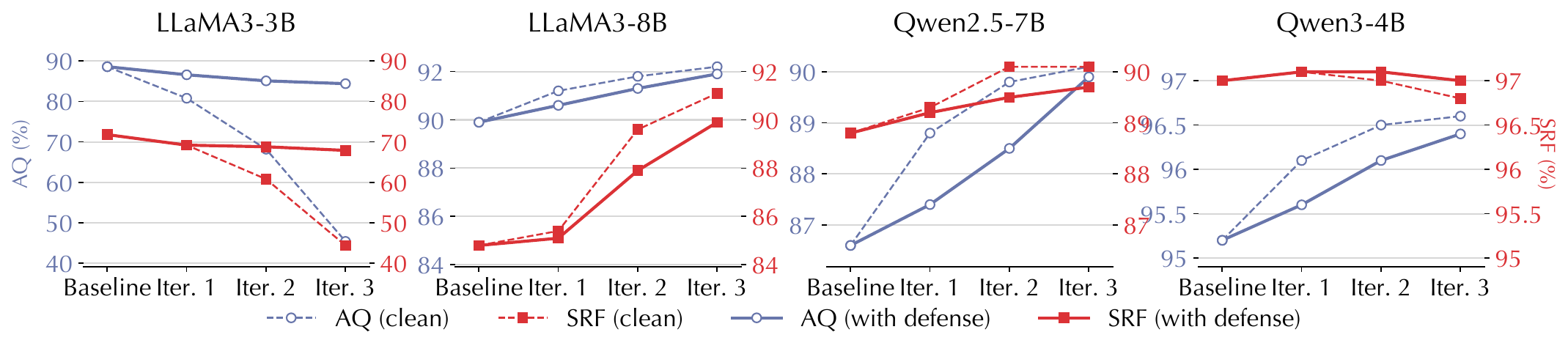}}
\caption{\small SRO iterations under attack-free conditions, with defense (solid) and without defense (dashed)}
\label{fig/exp_results_defense_clean}
\end{center}
\end{figure}

\paragraph{Performance without an attack.}
\cref{fig/exp_results_defense_clean} shows that the defense largely retains the gains of clean SRO for the three stronger models. After three iterations, defended AQ reaches 91.9\%, 89.9\%, and 96.4\% for {\tt Llama-3.1-8B}, {\tt Qwen-2.5-7B}, and {\tt Qwen-3-4B}, respectively, only 0.2--0.3 percentage points below undefended training. For {\tt Qwen-2.5-7B}, AQ starts at 86.6\% and reaches 88.8\%, 89.8\%, and 90.1\% without defense, versus 87.4\%, 88.5\%, and 89.9\% with defense over three iterations. As undefended AQ approaches saturation, the defense continues improving and nearly matches its final AQ despite a smaller initial gain. For {\tt Llama-3.2-3B}, the defense raises final AQ from 45.4\% to 84.4\% and SRF from 44.4\% to 67.9\%, though both remain below their initial values of 88.6\% and 71.8\%, respectively. These results suggest that the defense can also mitigate degradation during clean SRO while largely preserving gains in stable models. The benefit depends on evaluator drift and the cost of suppressing useful updates; \msec{app:defense_alpha}, \msec{app:defense_removed}, and \msec{app:defense_adaptive} examine this trade-off and an adaptive extension.

\begin{figure}[!ht]
\centering
\includegraphics[width=\columnwidth]{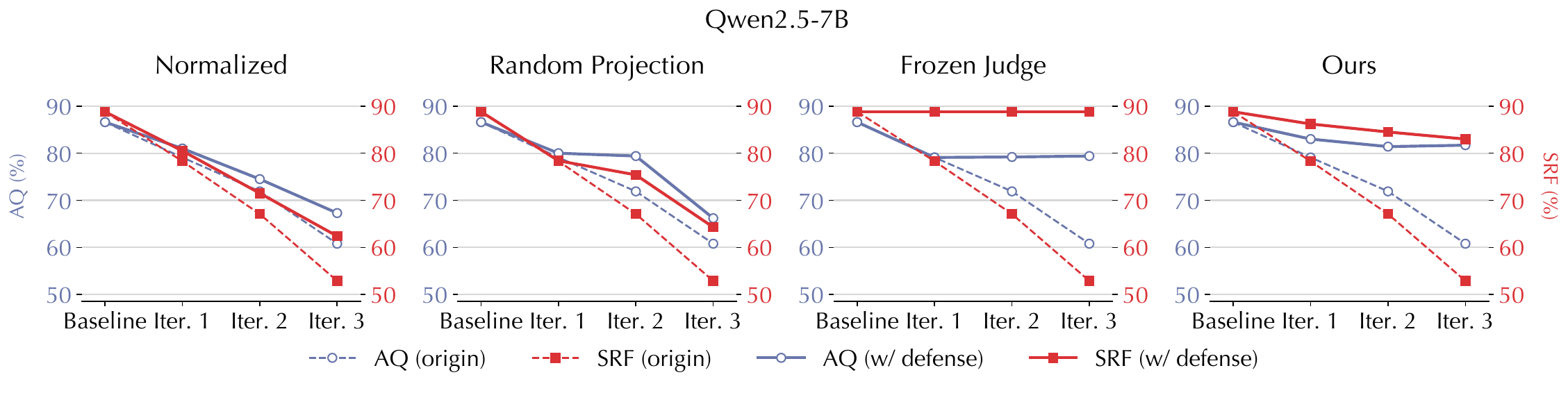}
\caption{\small Defense ablations on {\tt Qwen-2.5-7B} under Re-Ranking with SRLM.}
\label{fig:defense_ablation_attack}
\end{figure}

\paragraph{Defense ablations.}
We compare our defense with three baselines: normalized updates, locally norm-matched random projection, and Frozen Judge under Re-Ranking with SRLM (\cref{fig:defense_ablation_attack}). On {\tt Qwen-2.5-7B}, our defense retains the highest AQ across all three iterations. Frozen Judge, the strongest baseline at iteration~3, drops sharply from 86.6\% to 79.1\% AQ at iteration~1 and recovers only slightly to 79.4\% by iteration~3, remaining below our defense throughout training. Our defense finishes at 81.7\% AQ, compared with 67.3\% for normalization and 66.2\% for random projection. Control definitions and additional results are in Appendix~\ref{app:defense_ablation}.

\section{Conclusion}
\label{sec:conclud}

This paper exposes the fragility of self-rewarding optimization (SRO) via Re-Ranking and Preamble attacks that manipulate only the training-question distribution yet drive iterative collapse. We then analyze the failure mechanism, identifying self-reward fidelity (SRF), the on-policy agreement between self-evaluated and oracle rewards, as a diagnostic of self-supervision reliability, and showing that shared answer/reward parameters amplify low-fidelity updates. This analysis motivates Projection Decoupling Defense, which uses verdict-score gradients to identify and attenuate SRO updates along directions that affect self-evaluation, substantially improving robustness under adversarial conditions. Our work opens three directions for future research: \mct{i} developing adaptive defenses, \mct{ii} understanding how SRF dynamics scale with model capability, and \mct{iii} exploring whether similar phase-transition regions arise in other forms of self-supervised training.

\bibliography{iclr2027_conference}
\bibliographystyle{iclr2025_conference}

\appendix

\input{appendix}

\end{document}

%% file: prestuff.tex
\usepackage{epsfig,amsmath,amsfonts,epsfig,multirow,makecell,caption,soul,csquotes,color,wrapfig,subcaption,mathtools,bm,spverbatim,booktabs,tcolorbox,diagbox}
\usepackage[e]{esvect}

\makeatletter
\newif\if@restonecol
\makeatother

\usepackage[boxed, ruled, vlined, linesnumbered]{algorithm2e}
\SetKwRepeat{Do}{do}{while}

\newenvironment{changemargin}[2]{\begin{list}{}{
	\setlength{\topsep}{0pt}\setlength{\leftmargin}{0pt}
	\setlength{\rightmargin}{0pt}
	\setlength{\listparindent}{\parindent}
	\setlength{\itemindent}{\parindent}
	\setlength{\parsep}{0pt plus 1pt}
	\addtolength{\leftmargin}{#1}\addtolength{\rightmargin}{#2}
	}\item}
	{\end{list}}

\usepackage[first=0,last=9]{lcg}
\usepackage{colortbl}
\definecolor{Gray}{gray}{0.8}
\colorlet{Red}{red!10!white}
\colorlet{Blue}{blue!10!white}

\usepackage{hyperref}

\newcommand{\msec}[1]{\S\ref{#1}}
\newcommand{\mref}[1]{\,\ref{#1}}
\newcommand{\meq}[1]{Eq.\,\ref{#1}}
\newcommand{\mcite}[1]{\,\cite{#1}}

\newcommand{\mct}[1]{(#1)}

\tcbuselibrary{breakable}
\newtcolorbox{mtbox}[1]{breakable, left=0.25mm, right=0.25mm, top=0.25mm, bottom=0.25mm, sharp corners, colframe=red!50!black, boxrule=0.5pt, title={#1}, fonttitle=\bfseries, coltitle=red!50!black, attach title to upper={\ --\ }}

\usepackage{scalerel}[2016/12/29]

\makeatletter
\providecommand{\leadsfrom}{%
  \mathrel{\mathpalette\reflect@squig\relax}%
}
\newcommand{\reflect@squig}[2]{%
  \reflectbox{$\m@th#1\leadsto$}%
}
\makeatother

\def\eqref#1{equation~\ref{#1}}

\def\1{\bm{1}}

\DeclareMathAlphabet{\mathsfit}{\encodingdefault}{\sfdefault}{m}{sl}
\SetMathAlphabet{\mathsfit}{bold}{\encodingdefault}{\sfdefault}{bx}{n}

\def\gD{{\mathcal{D}}}

\def\gI{{\mathcal{I}}}

\def\gM{{\mathcal{M}}}

\def\gU{{\mathcal{U}}}

\def\sE{{\mathbb{E}}}

\def\sP{{\mathbb{P}}}

%% file: analysis.tex
We now analyze SRO collapse by separating the two roles played by the shared model: the answer generator $\pi_\theta$ and the reward generator $r_\theta$.
Let $L_\mathrm{AQ}$ be the answer-quality loss and $L_\mathrm{SRF}$ a differentiable surrogate loss for self-reward fidelity, constructed from oracle-labeled reward-generation preferences. Details of these two losses are defered to \msec{app:ans_re_opt}. Since SRF is an on-policy agreement metric evaluated on samples from $\pi_\theta$, whereas $L_\mathrm{SRF}$ is a training surrogate, we interpret local increases in $L_\mathrm{SRF}$ as local degradation of the reward generator's fidelity. We have the following proposition (detailed proofs and analysis in \msec{app:analysis_detail}).

\begin{proposition}[Transfer under Aligned Gradients]
\label{prop:clean_transfer}
Assume $L_\mathrm{SRF}$ is differentiable at $\theta_t$ and consider one model update $\theta_{t+1} = \theta_t - \eta \nabla_\theta L_\mathrm{AQ}$. If the answer-quality and self-reward gradients are positively aligned, $\langle \nabla_\theta L_\mathrm{AQ}, \nabla_\theta L_\mathrm{SRF} \rangle > 0$,
then there exists $\delta > 0$ such that, for every step size $\eta \in (0, \delta)$,
$L_\mathrm{SRF}(\theta_{t+1}) < L_\mathrm{SRF}(\theta_t)$.
\end{proposition}

Through this fixed-batch, local gradient-transfer view, we observe that shared parameters couple the two roles: when $\nabla_\theta  L_\mathrm{AQ}$ and $\nabla_\theta L_\mathrm{SRF}$ are positively aligned, an update that improves answer quality simultaneously reduces the SRF surrogate loss. This explains why SRO can initially improve both AQ and SRF when self-rewards remain reliable.

The same coupling, however, creates a conditional failure mode under noisy or adversarial self-rewards. Low SRF indicates that self-reward labels disagree more frequently with oracle labels, but does not by itself imply that the noisy AQ gradient reverses direction. However, if unreliable rewards corrupt the AQ preference signal sufficiently to make the noisy update locally misaligned with the SRF descent direction, shared parameters transfer this corruption into the reward-generator role. Formally:
\begin{proposition}[Transfer under Misaligned Gradients]
\label{prop:noisy_transfer}
Let $\tilde{L}_\mathrm{AQ}=L_\mathrm{AQ}+\Delta_\mathrm{AQ}$ be the noisy AQ objective induced by unreliable self-rewards, where $\Delta_{AQ}$ captures the corruption. Consider one update $\theta_{t+1} = \theta_t - \eta \nabla_\theta \tilde{L}_\mathrm{AQ}$.
Assume $L_\mathrm{SRF}$ has $\beta_\mathrm{SRF}$-Lipschitz gradient in a neighborhood of $\theta_t$. If the noisy AQ gradient misaligns with the SRF gradient, $\langle \nabla_\theta L_\mathrm{SRF}, \nabla_\theta \tilde{L}_\mathrm{AQ}\rangle \leq -\gamma$ for some $\gamma > 0$, then for all $0 < \eta <
    \frac{2\gamma}{\beta_\mathrm{SRF}\| \nabla_\theta \tilde{L}_\mathrm{AQ}\|^2}$, $L_\mathrm{SRF}(\tilde{\theta}_{t+1}) > L_\mathrm{SRF}(\theta_t)$.
\end{proposition}
This result exposes a one-step mechanism for iterative collapse: low-fidelity rewards corrupt the answer update, which in turn degrades the reward generator and renders future self-evaluations less reliable. A complete multi-step collapse theorem would further require linking SRF to gradient corruption across rounds and analyzing the resulting recurrence, which we leave for future work.

%% file: appendix.tex
\section{Proofs}
\label{sec:proof}

\input{proof}

\cref{prop:clean_transfer} explains why improvements in answer generation can coincide with improvements in the SRF surrogate when the two roles are aligned.
\cref{prop:noisy_transfer} gives the corresponding conditional failure mode: once low-fidelity self-rewards have corrupted the AQ preference signal enough to reverse the ideal AQ descent direction, the shared, positively aligned parameters can transfer that corrupted update to the SRF surrogate.
These propositions establish a one-step local mechanism; a full iterative collapse theorem would further require a relationship between SRF and preference-gradient corruption across rounds.

\section{Additional Experiments}

\subsection{Phase transition region of SRO on smaller model ({\tt LlaMA-3.2-3B})}
\label{sec:pt_llama3b}

\begin{figure}[ht]
\begin{center}
\centerline{\includegraphics[width=\columnwidth]{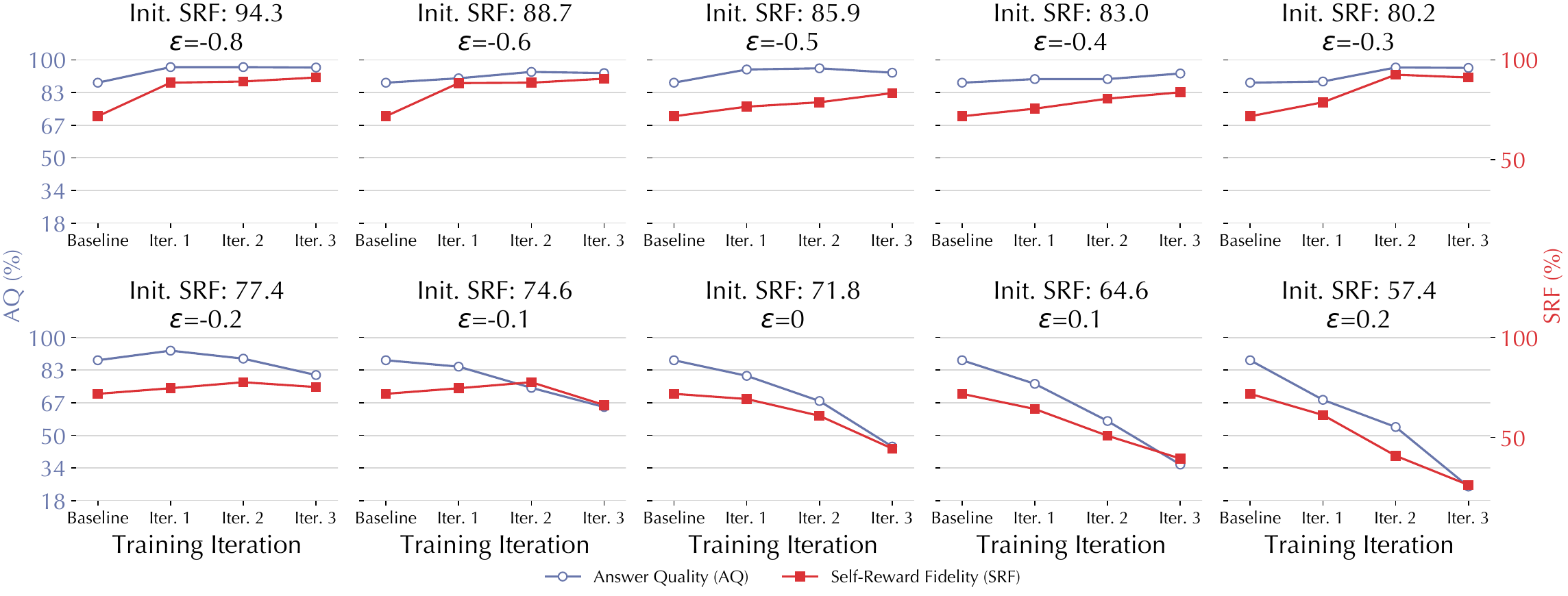}}
\caption{Phase-transition region in Llama-3.2-3B when manipulating the SRF.}
\label{fig/iteration_judge_noise_combined_dual_yaxis_llama3_3b_bold}
\end{center}
\end{figure}

\begin{figure}
\begin{center}
\centerline{\includegraphics[width=\columnwidth]{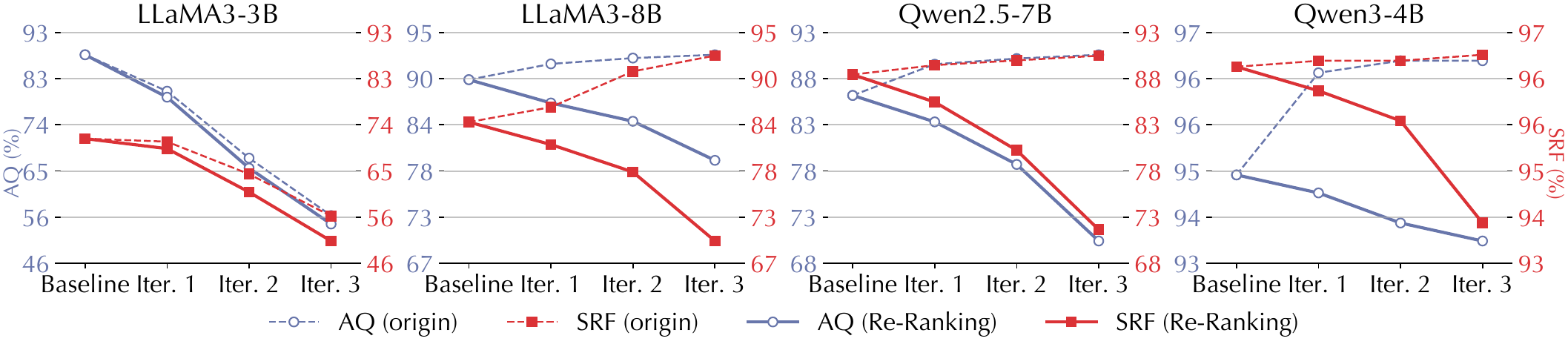}}
\caption{\small Safety alignment collapse with Re-Ranking attack under CREAM.}
\label{fig/exp_results_cream_re_ranking_safty}
\end{center}
\end{figure}

\begin{figure}[!h]
\begin{center}
\centerline{\includegraphics[width=\columnwidth]{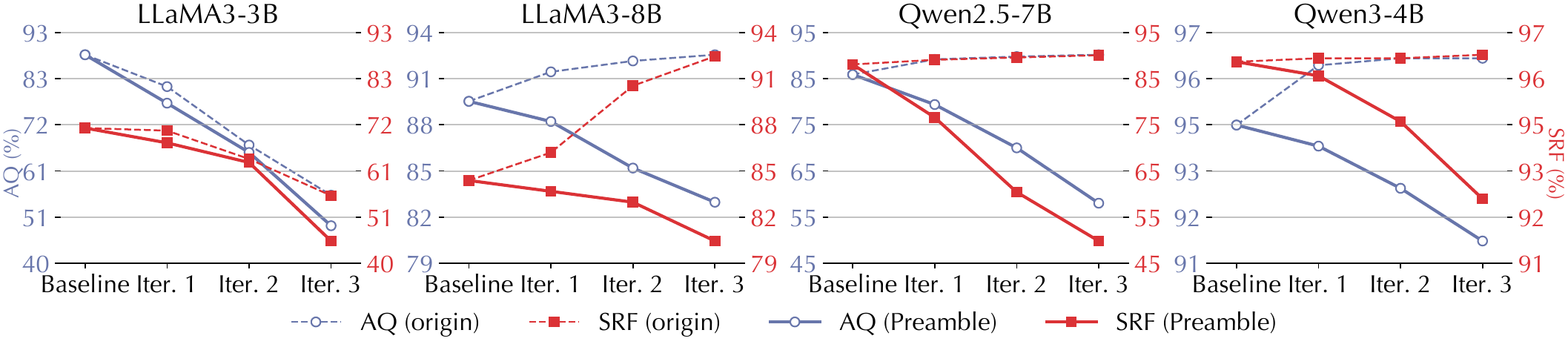}}
\caption{\small Safety alignment collapse with Preamble attack under CREAM.}
\label{fig/exp_results_cream_preamble_safty}
\end{center}
\end{figure}

Figure\mref{fig/transition_of_noise} establishes how the noise $\epsilon$ translates into a drop in SRF based on {\tt Llama3-8B}, revealing a transition from reliable to unreliable self-rewards as corruption increases. Figure\mref{fig/iteration_judge_noise_combined_dual_yaxis_llama3_3b_bold} then asks how that fidelity transition manifests in SRO for a less capable model, such as {\tt LlamA-3.2-3B}. We also observe a phase-transition region in {\tt Llama-3.2-3B} as SRF is manipulated by noise $\epsilon$, occurring at a different noise level than for {\tt Llama-3.1-8B}. In the $\epsilon\le$ -0.3 regime, one to two rounds of iteration of SRO yield consistent improvements in both AQ and SRF, followed by saturation. Near the transition region (Init.\ SRF $\approx 74.6$--$77.4$, $\epsilon\in\{-0.2,-0.1\}$), SRF can improve early, but AQ ceases to improve and may regress with additional rounds, indicating instability when optimizing against partially corrupted self-rewards. As Init.\ SRF falls further (e.g., $\epsilon\ge 0$, Init.\ SRF $\le 71.8$), both AQ and SRF decrease monotonically across iterations, demonstrating compounding error under low-fidelity self-reward. Taken together with Figure\mref{fig/transition_of_noise}, these results suggest that ,in these settings, the initial SRF set by noise is associated with whether iterative self-reward training leads to self-correction or collapse, and that this behavior persists when moving from {\tt Llama-3.1-8B} to {\tt Llama-3.2-3B}.

\subsection{Attack performance}
\label{app:attack}

{\bf SRO Collapse (CREAM).}
Compared with SRLM, CREAM achieves higher final AQ in the clean loop; for example, {\tt Llama-3.2-3B} ends at $56.4/56.3$ instead of $45.4/44.4$, where each pair denotes AQ/SRF in percent. Under the Re-Ranking attack (\cref{fig/exp_results_cream_re_ranking_safty}), the iteration~3 scores are $54.7/51.3$ for {\tt Llama-3.2-3B}, $80.2/70.5$ for {\tt Llama-3.1-8B}, $71.2/72.4$ for {\tt Qwen-2.5-7B}, and $94.1/94.4$ for {\tt Qwen-3-4B}. Under the Preamble attack (\cref{fig/exp_results_cream_preamble_safty}), the corresponding iteration~3 scores are $49.4/45.9$, $83.4/80.9$, $58.7/50.5$, and $91.9/93.1$, respectively. CREAM retains higher final AQ than SRLM in all eight model--attack settings, but its SRF is lower for {\tt Llama-3.1-8B} and {\tt Qwen-3-4B} under Re-Ranking. Both attacks reduce AQ and SRF relative to clean CREAM, although {\tt Qwen-3-4B} remains comparatively robust.

\begin{figure}
\begin{center}
\centerline{\includegraphics[width=\columnwidth]{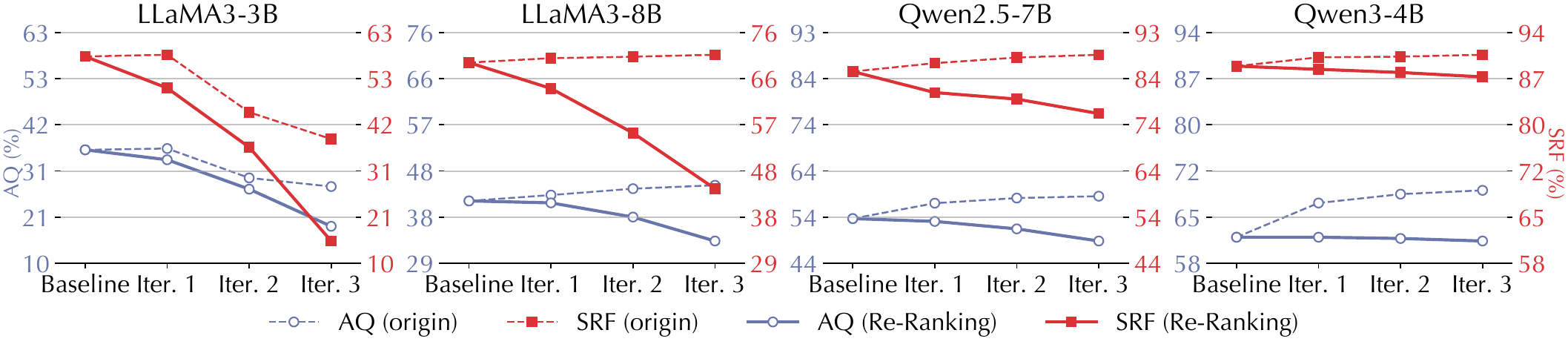}}
\caption{\small General knowledge collapse with Re-Ranking Attack, using MMLU-Pro and SRO (CREAM).}
\label{fig/exp_results_cream_re_ranking_mmlu}
\end{center}
\end{figure}

\begin{figure}
\begin{center}
\centerline{\includegraphics[width=\columnwidth]{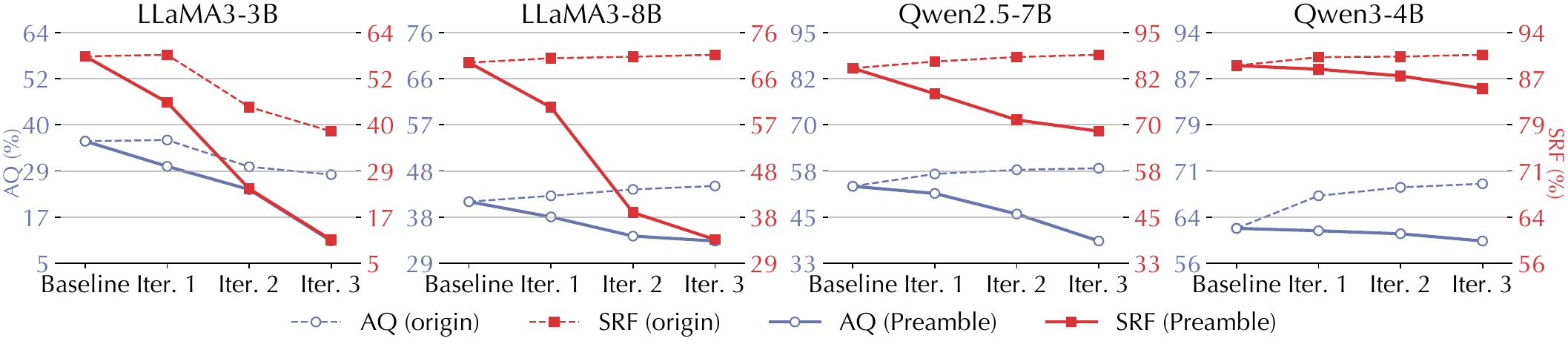}}
\caption{\small General knowledge collapses with Preamble Attack, using MMLU-Pro and SRO (CREAM).}
\label{fig/exp_results_cream_preamble_mmlu}
\end{center}
\end{figure}

\cref{fig/exp_results_cream_re_ranking_mmlu,fig/exp_results_cream_preamble_mmlu} track AQ and SRF on MMLU-Pro across SRO (CREAM). In the clean loop (dashed), CREAM yields consistent gains for sufficiently capable models, such as {\tt Qwen3-4B}, which improves from $62.4/89.4$ to $69.8/91.2$. Under attacks, degradation compounds with iteration: Re-Ranking already drops {\tt Llama-3.2-3B} to $19.1/15.7$ at iteration 3, and Preamble is typically more destructive on MMLU-Pro, collapsing it further to $11.4/11.7$. Notably, for stronger models, the SRF can remain relatively high even as AQ deteriorates, for example, {\tt Qwen3-4B} at iteration 3: $61.8/87.7$ under Re-Ranking and $60.3/85.6$ under Preamble.

\begin{figure}
\begin{center}
\centerline{\includegraphics[width=\columnwidth]{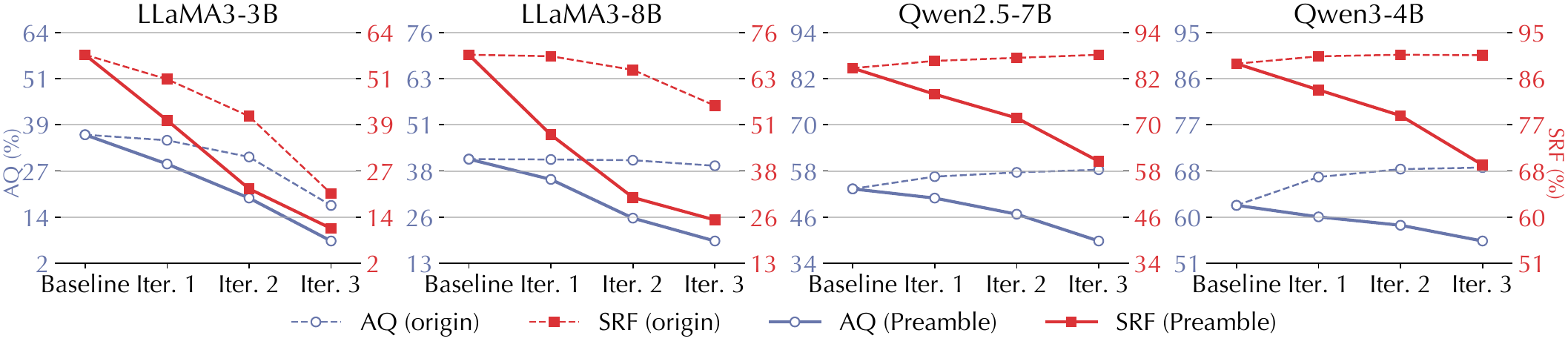}}
\caption{\small General knowledge collapses with Preamble Attack, using MMLU-Pro and SRO (SRLM).}
\label{fig/exp_results_srlm_preamble_mmlu}
\end{center}
\end{figure}

\begin{figure}
\begin{center}
\centerline{\includegraphics[width=\columnwidth]{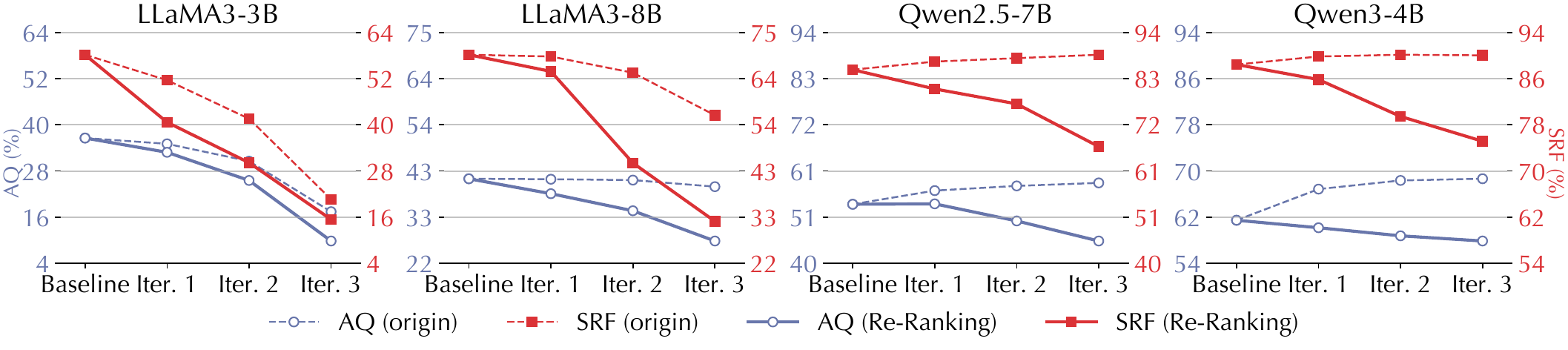}}
\caption{\small General knowledge collapse with Re-Ranking Attack, using MMLU-Pro and SRO (SRLM).}
\label{fig/exp_results_srlm_re_ranking_mmlu}
\end{center}
\end{figure}

\begin{figure}
\begin{center}
\centerline{\includegraphics[width=\columnwidth]{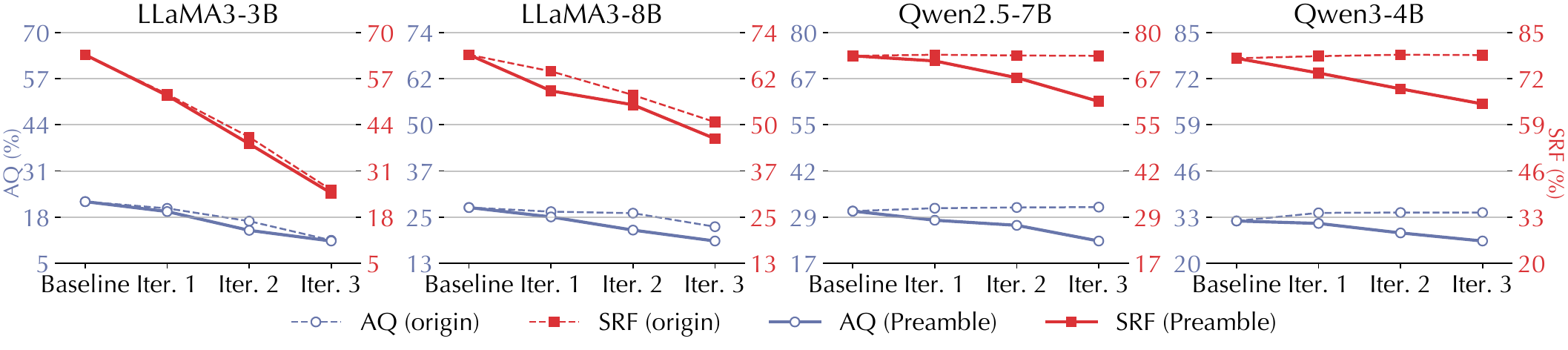}}
\caption{\small Logic reasoning collapses with Preamble Attack, using ARC-challenge and SRO (SRLM).}
\label{fig/exp_results_srlm_preamble_arc}
\end{center}
\end{figure}

\begin{figure}
\begin{center}
\centerline{\includegraphics[width=\columnwidth]{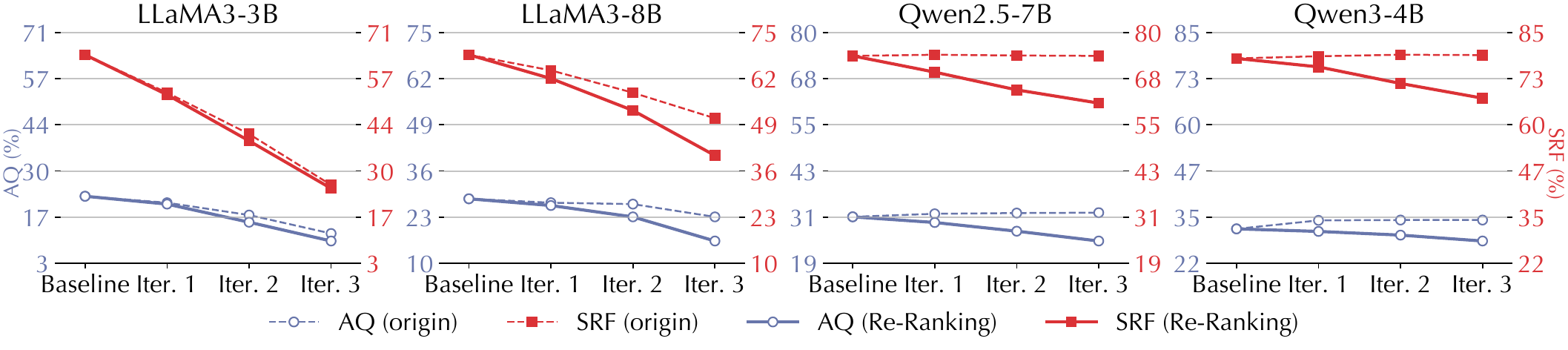}}
\caption{\small Logic reasoning collapses with Re-Ranking Attack, using the ARC-challenge and SRO (SRLM).}
\label{fig/exp_results_srlm_re_ranking_arc}
\end{center}
\end{figure}

{\bf SRO Collapse on Other Tasks.}
Beyond safety alignment, we test whether manipulating SRF spills over to downstream capability on general knowledge and reasoning benchmarks. \cref{fig/exp_results_srlm_re_ranking_mmlu,fig/exp_results_srlm_preamble_mmlu} show the performance on MMLU-Pro (general knowledge) and \cref{fig/exp_results_srlm_re_ranking_arc,fig/exp_results_srlm_preamble_arc} show ARC-Challenge (reasoning benchmark).

On MMLU-Pro (general knowledge), clean SRLM is largely capability-preserving for {\tt Qwen} and can even improve knowledge accuracy, for example, {\tt Qwen3-4B} increases from $62.4/89.4$ to $69.6/91.0$ by Iteration~3. However, both attacks induce a monotonic collapse across iterations. The same model drops to $58.8/76.1$ (Re-Ranking) or $55.6/70.1$ (Preamble) at Iteration~3, which erases the clean gains and lowers SRF by roughly $15$ to $21$ points relative to the clean trajectory. {\tt Llama} models are substantially more brittle. Even without attacks SRF deteriorates with additional self-rewarding steps, and under attack, the model collapses. For example, {\tt Llama-3-8B} falls from $41.9/70.2$ to $19.7/25.4$ with Preamble at Iteration~3. Overall, SRF manipulation does not stay confined to ``safety behavior''. Once SRLM trains on a compromised reward signal, it also degrades generalizable knowledge competence.

On ARC-Challenge (reasoning benchmark), the coupling persists but is typically milder for {\tt Qwen}: {\tt Qwen3-4B} is nearly stable in the clean run, performing $32.2/78.5$ and increasing to $34.6/79.4$, yet drops to $28.9/67.7$ (Re-Ranking) and $26.5/65.5$ (Preamble) at Iteration~3 (about $8$ AQ points and $12$ to $14$ SRF points below clean). {\tt Llama} again exhibits larger degradation. {\tt Llama3-8B} reaches $16.3/40.4$ under Re-Ranking. Taken together, the attacked curves show a clear iterative amplification effect. Each additional SRLM iteration compounds the error, suggesting SRF is tightly entangled with downstream competence rather than an isolated alignment knob.

\begin{figure}
\begin{center}
\centerline{\includegraphics[width=0.7\columnwidth]{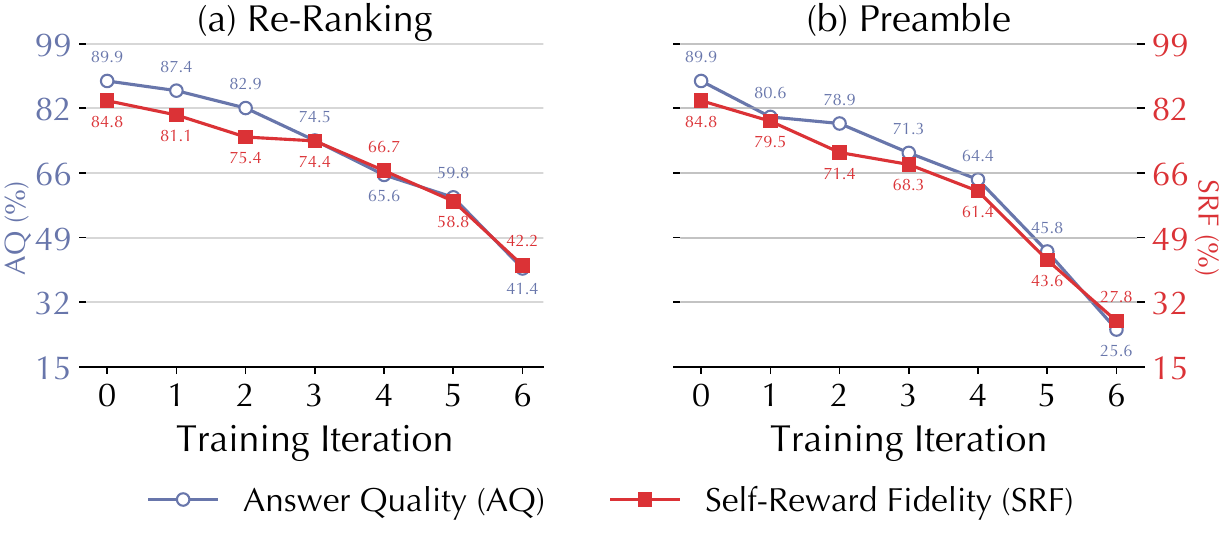}}
\caption{\small The effect of the number of iterations.}
\label{fig/num_iter}
\end{center}
\end{figure}

{\bf Attack Ablation.}
We further extend the training duration to $6$ iterations to observe the long-term dynamics of the collapse under both the Re-Ranking and Preamble triggers. \cref{fig/num_iter} plots the progression of AQ and SRF for {\tt Llama3-8B}.

In both scenarios, the performance degradation is continuous, indicating that the model fails to self-correct over prolonged training. The Preamble attack (right) induces a more rapid and severe decline compared to the Re-Ranking attack (left). By iteration~6, AQ falls to $25.6\%$ under Preamble and $41.4\%$ under Re-Ranking; the corresponding SRF values are $27.8\%$ and $42.2\%$, respectively. Furthermore, we observe that the gap between AQ and SRF is significantly reduced in later iterations. This convergence suggests that as the model collapses, its ability to distinguish between high-quality answers and its own degenerated outputs vanishes.

\subsection{Defense Performance on CREAM}
\label{app:defense_cream}

\begin{figure}[!ht]
\begin{center}
\centerline{\includegraphics[width=\columnwidth]{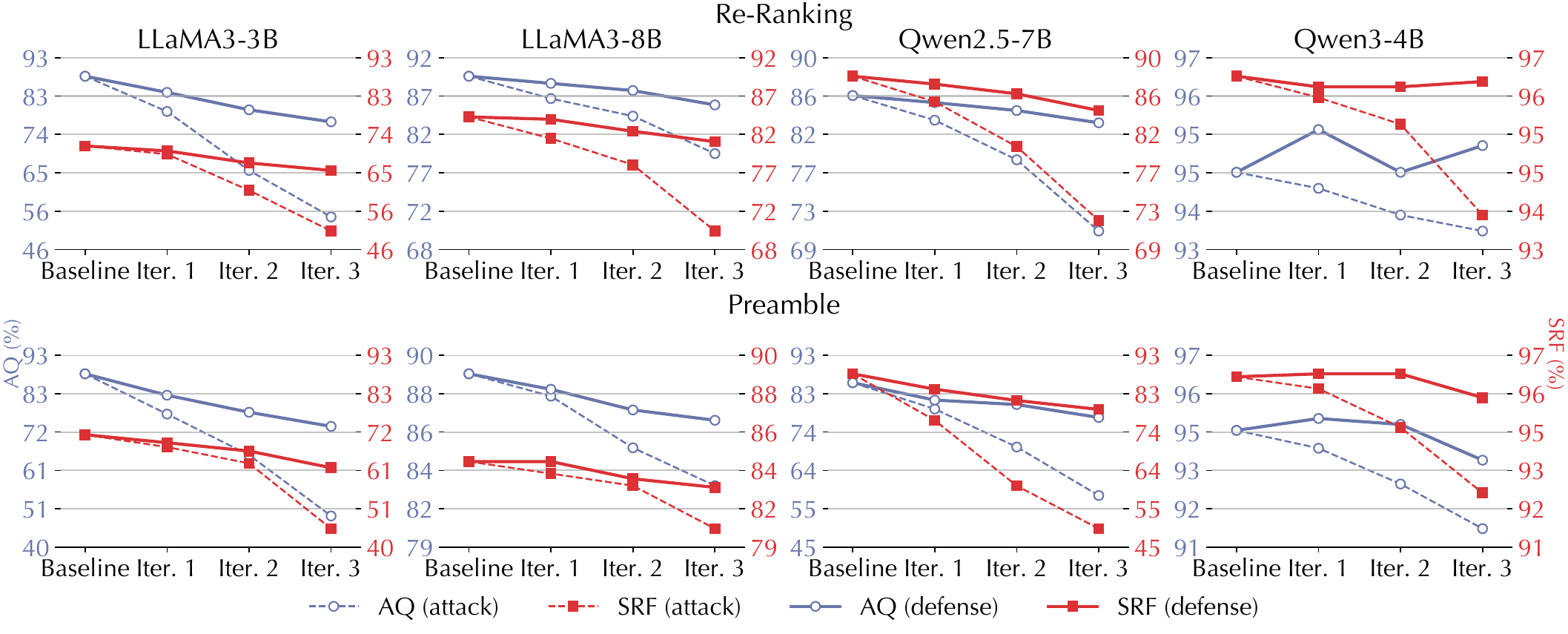}}
\caption{\small Defense against two proposed attacks on CREAM (upper: ReRanking attack; lower: Preamble attack).}
\label{fig/exp_results_defense_cream}
\end{center}
\end{figure}

We evaluate Projection Decoupling Defense with CREAM on safety alignment across four models under Re-Ranking and Preamble attacks, using three SRO iterations, $N_{\mathcal{C}}=16$ probes, and $\alpha=0.5$. \cref{fig/exp_results_defense_cream} shows AQ improvements of 1.6--24.7 percentage points over undefended CREAM at iteration~3. Under Preamble, the defense raises {\tt Qwen-2.5-7B}'s AQ from 58.7\% to 78.0\% and SRF from 50.5\% to 80.0\%.

\subsection{Estimating SRF with Limited Oracle Labels}
\label{app:srf_limited_labels}

\paragraph{Experimental setup.}
SRF (\meq{eq:srq}) requires oracle labels on answers sampled from the current policy. We compare full-evaluation SRF with estimates using $32$--$256$ oracle-labeled responses per checkpoint, for the iteration-3 checkpoints of {\tt Llama-3.1-8B} with SRLM on SafeRLHF, without defense, under clean training and Re-Ranking. Labels are provided by humans or a stronger evaluator.

\begin{table}[!ht]
\centering
\small
\caption{\small SRF estimated from a limited oracle budget on {\tt Llama-3.1-8B} with SRLM on SafeRLHF (iteration~3, no defense). Entries are mean SRF (\%) with standard deviations in parentheses.}
\label{tab:srf_limited_labels}
\begin{tabular}{lrr}
\toprule
Oracle labels & Clean SRF (\%) & Re-Ranking SRF (\%) \\
\midrule
Full evaluation & 91.1 & 74.4 \\
32 & 90.5 (3.0) & 73.3 (4.3) \\
64 & 90.7 (2.2) & 73.6 (2.9) \\
128 & 90.9 (1.3) & 74.6 (2.3) \\
256 & 91.2 (1.1) & 74.4 (1.7) \\
\bottomrule
\end{tabular}
\end{table}

\paragraph{Results.}
As shown in \cref{tab:srf_limited_labels}, with 256 labels, the mean estimated SRF is within 0.1\% of the full-evaluation reference. Even with only 32 labels, the estimates still separate the clean and attack conditions. These results suggest that SRF can be applied in practice through low-budget labeling.

\section{Defense Setting}

\subsection{Verdict-Score Definition}
\label{app:defense_score}

For a reward generator that produces binary verdicts under a fixed evaluation prompt, we define the \emph{verdict score} used by Projection Decoupling Defense (\msec{sec:defense}) for each fixed probe $(x_i,y_i)\in\mathcal{C}$ as
\begin{equation}
\ell_i(\theta)=
\log r_\theta(c_+\mid x_i,y_i)
-\log r_\theta(c_-\mid x_i,y_i),
\label{eq:fj-score}
\end{equation}
where $c_+$ and $c_-$ are the positive and negative verdict sequences, respectively, and $r_\theta(c\mid x,y)$ denotes the probability of verdict sequence $c$ under that prompt. Sequence log probabilities sum over all tokens, including termination tokens. For example, in the safety alignment task, $c_+$ and $c_-$ correspond to \texttt{SAFE} and \texttt{UNSAFE}. A positive verdict score favors the positive verdict, a negative score favors the negative verdict, and zero indicates equal probabilities for the two verdict sequences. Its gradient measures the sensitivity of this relative preference to parameter changes. Here, a probe is a fixed question--answer pair; the verdict score is the scalar computed on that pair. Constructing verdict scores requires no oracle labels, whereas SRF measures agreement with oracle rewards on answers sampled from the current policy.

\subsection{Verdict-Score Basis}
\label{app:defense_basis}

The verdict-score gradient subspace in \meq{eq:defense-verdict-gradients} can equivalently be represented by stacking the round-start gradients into a verdict-score Jacobian:
\begin{equation}
J=\begin{bmatrix}g_1^\top\\ \vdots\\ g_{N_{\mathcal{C}}}^\top\end{bmatrix}
\in\mathbb{R}^{N_{\mathcal{C}}\times d}.
\label{eq:fj-jacobian}
\end{equation}
Its row space is exactly $\mathcal{S}$. For an orthonormal basis $Q$ spanning this space, $JQQ^\top=J$. Thus, for a candidate displacement $\Delta$ and its defended counterpart $\widehat\Delta=\Delta-\alpha QQ^\top\Delta$, we have $J\widehat\Delta=(1-\alpha)J\Delta$. The gradients are evaluated once per SRO round, and the resulting basis is reused for every optimizer step within that round. The matrix $J$ need not be retained after constructing the basis.

We construct the basis using the small Gram matrix $G=JJ^\top\in\mathbb{R}^{N_{\mathcal{C}}\times N_{\mathcal{C}}}$. Let $U$ contain orthonormal eigenvectors of $G$ corresponding to its positive eigenvalues, and let $\Lambda$ be the diagonal matrix of those eigenvalues. In exact arithmetic,
\begin{equation}
Q=J^\top U\Lambda^{-1/2}
\label{eq:defense-gram-basis}
\end{equation}
has orthonormal columns spanning $\mathcal{S}$. If numerical truncation retains only a subset of these directions, let $\widetilde Q$ denote the retained orthonormal basis and $\widetilde P=\widetilde Q\widetilde Q^\top$ its projector. The corresponding defended displacement $\widehat\Delta=\Delta-\alpha\widetilde P\Delta$ satisfies
\begin{equation}
J\widehat\Delta=(1-\alpha)J\Delta
+\alpha J(I-\widetilde P)\Delta.
\label{eq:defense-truncation-residual}
\end{equation}
The residual vanishes when the retained basis spans all verdict-score gradients; otherwise, it captures the contribution from omitted directions. This basis approximation is separate from the error of using round-start verdict-score gradients to predict actual verdict-score changes later in the round.

\section{Defense Ablation}
\label{app:defense_ablation}
\label{app:defense_normalization}

We examine whether the benefits of Projection Decoupling Defense arise from reducing update magnitude, selecting directions identified by verdict-score gradients, or constraining judge adaptation.

\subsection{Ablation Methods}
\label{app:defense_ablation_methods}

We compare Projection Decoupling Defense (\msec{sec:defense}) with standard SRO and three controls. Below, $\Delta_s^{\mathrm{raw}}$ denotes the ordinary DPO/Adam displacement at each control's current parameters. Both normalization and projection operate on this displacement after Adam, rather than on raw gradients.

\paragraph{Normalized updates.}
We record the defended run's actual update norm $m_s=\|\widehat\Delta_s^{\mathrm{def}}\|_2$ at every optimizer step, separately for each training condition. From the same initialization, the control replays these norms while preserving its own ordinary update direction:
\begin{equation}
\Delta_s^{\mathrm{norm}}
=\frac{m_s}{\|\Delta_s^{\mathrm{raw}}\|_2}\Delta_s^{\mathrm{raw}}.
\label{eq:defense_norm_control}
\end{equation}
For nonzero $\Delta_s^{\mathrm{raw}}$, this matches the defended run's step magnitude even as the parameter trajectories and subsequent preferences diverge.

\paragraph{Random projection.}
At the start of each SRO round $t$, we compute the verdict-score Jacobian basis $Q_t^{\mathrm{verdict}}$ on the fixed probes at the control's round-start parameters, using the same basis construction as the main defense. We orthogonalize a Gaussian matrix to obtain $Q_t^{\mathrm{rand}}$ with the same number of columns as $Q_t^{\mathrm{verdict}}$. Both bases remain fixed throughout the round, matching the main defense's update schedule. At each optimizer step $s$ within round $t$, we form
\begin{equation}
\begin{aligned}
u_s^{\mathrm{verdict}} &= \Delta_s^{\mathrm{raw}}-\alpha Q_t^{\mathrm{verdict}}(Q_t^{\mathrm{verdict}})^\top\Delta_s^{\mathrm{raw}}, \\
u_s^{\mathrm{rand}} &= \Delta_s^{\mathrm{raw}}-\alpha Q_t^{\mathrm{rand}}(Q_t^{\mathrm{rand}})^\top\Delta_s^{\mathrm{raw}}.
\end{aligned}
\label{eq:defense_random_control}
\end{equation}
We use $u_s^{\mathrm{verdict}}$ only to determine the target norm and apply
\begin{equation}
\Delta_s^{\mathrm{rand}}=
\frac{\|u_s^{\mathrm{verdict}}\|_2}{\|u_s^{\mathrm{rand}}\|_2}u_s^{\mathrm{rand}},
\qquad \|u_s^{\mathrm{rand}}\|_2>0.
\label{eq:defense_random_norm_control}
\end{equation}
This per-step norm matching preserves the randomly filtered direction while matching the norm obtained by applying the fixed round-start verdict-score gradient basis to the same current displacement. The target norm varies with $\Delta_s^{\mathrm{raw}}$; the verdict-score Jacobian and both bases are refreshed only at the start of the next SRO round. Unlike norm replay, this does not ensure equal step magnitudes across the two complete training runs.

\paragraph{Frozen Judge.}
We fix the judge to the initial checkpoint, $\phi=\theta_0$, while updating the answer policy $\pi_{\theta_t}$ with DPO. Each iteration, the current policy generates new answers and $r_\phi$ scores them to construct preference pairs. We also recompute on-policy SRF using $r_\phi$ on newly generated answers, following \meq{eq:srq} with $r_\theta$ replaced by $r_\phi$. Freezing this evaluator is distinct from freezing the reference model in the DPO loss.

\subsection{Ablation Results}
\label{app:defense_ablation_results}

We evaluate SRLM on safety alignment for three SRO iterations with $N_{\mathcal{C}}=16$ and $\alpha=0.5$. We compare all controls on {\tt Llama-3.1-8B} under Re-Ranking and clean training.

\begin{figure}[!ht]
\centering
\includegraphics[width=\columnwidth]{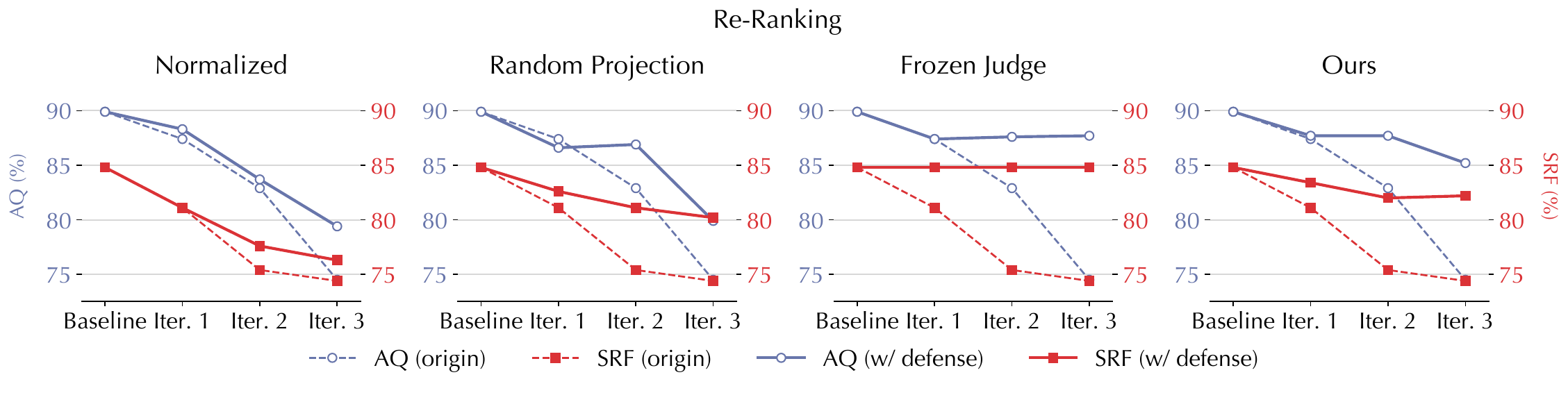}
\par\medskip
\includegraphics[width=\columnwidth]{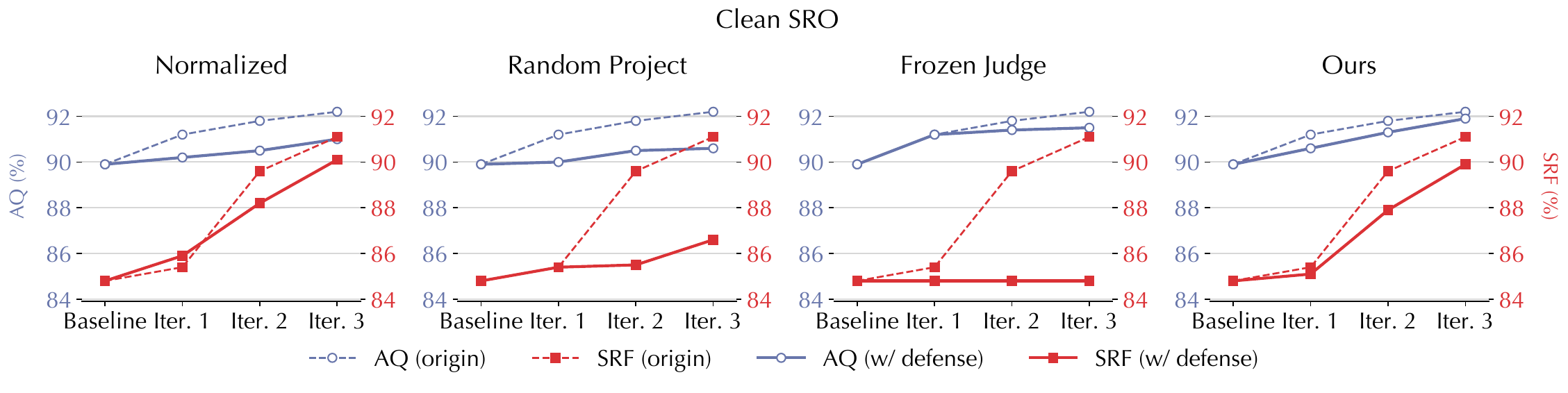}
\caption{\small Defense ablations on {\tt Llama-3.1-8B} with SRLM for safety alignment, under Re-Ranking (top) and without an attack (bottom). In each panel, solid curves show the named method and dashed curves repeat standard SRO under the same condition. Blue circles denote AQ (left axis), and red squares denote SRF (right axis), both in percent. Baseline is the shared initialization; axis ranges differ between attack and clean training. Random Project uses local norm matching with a random basis refreshed each SRO iteration; Normalized replays defended-run norms. Frozen Judge SRF is recomputed on newly generated answers at each iteration.}
\label{fig:defense_ablation_controls}
\end{figure}

\paragraph{Robustness under attack.}
At iteration~3, our defense reaches $85.2\%$ AQ and $82.2\%$ SRF (\cref{fig:defense_ablation_controls}), exceeding standard SRO by $10.7/7.8$ percentage points, normalization by $5.8/5.9$ points, and locally norm-matched random projection by $5.3/2.0$ points (AQ/SRF). These results support the value of directions identified by verdict-score gradients beyond scalar update rescaling. Frozen Judge performs best under attack ($87.7\%/84.8\%$), although all methods finish below the initial AQ.

\paragraph{Adaptation without an attack.}
Our defense reaches $91.9\%$ AQ and $89.9\%$ SRF, with AQ only $0.3$ points below standard SRO and $0.4$ points above Frozen Judge. Its SRF exceeds Frozen Judge by $5.1$ points but trails normalization by $0.2$ points. Thus, freezing provides stronger attack robustness, while projection retains more improvement during clean training. Frozen Judge's constant $84.8\%$ SRF is an empirical result of on-policy reevaluation, not a guarantee of freezing. The reported point estimates do not establish statistical significance.

\subsection{Effect of Protection Strength}
\label{app:defense_alpha}

\begin{figure}[!ht]
\centering
\includegraphics[width=\columnwidth]{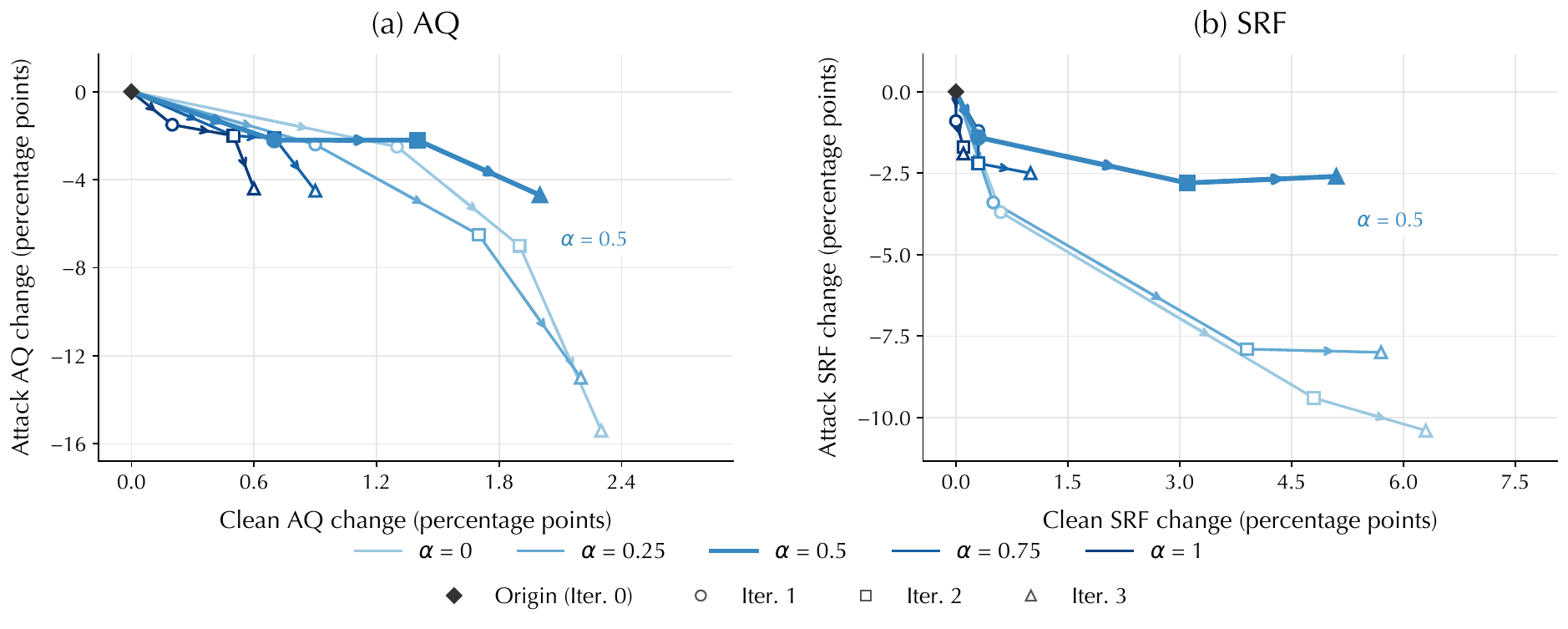}
\caption{\small Effect of protection strength $\alpha$ on {\tt Llama-3.1-8B} with SRLM for safety alignment. Panels show changes in (a) AQ and (b) SRF relative to the shared initialization ($89.9\%$ AQ and $84.8\%$ SRF), in percentage points. Each point pairs the change under clean training (horizontal axis) with the change under Re-Ranking (vertical axis) at the same iteration. Colors denote $\alpha$; circles, squares, and triangles denote iterations~1--3, and arrows indicate progression from initialization (black diamond). The highlighted $\alpha=0.5$ trajectory retains clean learning gains while substantially reducing degradation under attack relative to $\alpha=0$.}
\label{fig:defense_alpha}
\end{figure}

\paragraph{Experimental setup.}
We ablate the protection strength $\alpha$ in \meq{eq:fj-update} to examine the tradeoff between attack robustness and learning under clean training. We sweep $\alpha\in\{0,0.25,0.5,0.75,1\}$ on {\tt Llama-3.1-8B} with SRLM for safety alignment, under both Re-Ranking and attack-free training. All other settings are held fixed, following \msec{sec:attack} and \msec{sec:defense} and the control ablations above, including $N_{\mathcal{C}}=16$ probes. We measure AQ and SRF at initialization and after each of three SRO iterations, using the shared initialization and undefended SRO ($\alpha=0$) as reference points. \cref{fig:defense_alpha} pairs the changes from initialization under clean and attacked training for each iteration.

\paragraph{Results and parameter selection.}
Increasing $\alpha$ from $0.25$ to $0.5$ substantially improves robustness at a small cost to clean learning: at iteration~3, attacked AQ rises from $76.9\%$ to $85.2\%$ ($+8.3$ percentage points), while clean AQ decreases by only $0.2$ points. SRF follows the same pattern, with an attacked gain of $5.4$ points and a clean reduction of $0.6$ points. Beyond $\alpha=0.5$, the additional robustness gains diminish. Raising $\alpha$ to $0.75$ or $1$ improves attacked AQ by only $0.2$ or $0.3$ points, respectively, while reducing clean AQ by $1.1$ or $1.4$ points. The corresponding attacked SRF gains are $0.1$ and $0.7$ points, compared with clean SRF reductions of $4.1$ and $5.0$ points. At $\alpha=0.5$, clean AQ and SRF reach $91.9\%$ and $89.9\%$, improving over initialization by $2.0$ and $5.1$ points, respectively; clean AQ remains just $0.3$ points below undefended SRO. These results support $\alpha=0.5$ as a default that balances robustness and continued learning in this setting. Nevertheless, attacked AQ and SRF remain $4.7$ and $2.6$ points below initialization, so the observed benefit is mitigation of degradation rather than its elimination

\paragraph{When does projection help?}
Because $\alpha$ scales every update component in the protected subspace regardless of whether it improves or degrades reward fidelity (\meq{eq:fj-update}), projection can prevent harmful evaluator changes but can also limit useful learning. We therefore expect it to help when the benefit of stabilizing the evaluator outweighs this cost: under attack, or for models whose clean SRO run is itself unstable, such as {\tt Llama-3.2-3B}, whose final clean AQ rises from $45.4\%$ to $84.4\%$ with the defense (\cref{fig/exp_results_defense_clean}). When clean SRO is already stable, the cost is small, with final AQ $0.2$--$0.3$ points below undefended training for the other three models. \msec{app:defense_removed} quantifies how much of the removed update is beneficial or harmful.

\subsection{Effect of Probe Count}
\label{app:defense_probe_count}

\paragraph{Experimental setup.}
We examine how the probe count affects final answer quality and the computational overhead of Projection Decoupling Defense (\msec{sec:defense}) on {\tt Llama-3.1-8B} with SRLM for safety alignment under Re-Ranking. We vary $N_{\mathcal{C}}\in\{4,8,16,32,64,128\}$ with $\alpha=0.5$ and all other settings fixed; each probe is a fixed question--answer pair used by the defense throughout training. We report AQ after the third SRO iteration and the additional wall-clock time for probe computation per SRO iteration, measured in seconds.

\begin{figure}[!ht]
\centering
\includegraphics[width=0.75\columnwidth]{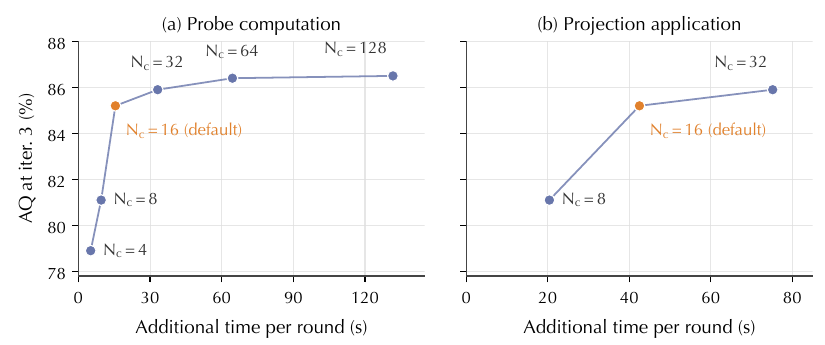}
\caption{\small Effect of probe count on {\tt Llama-3.1-8B} with SRLM for safety alignment under Re-Ranking, with $\alpha=0.5$. Each point shows AQ after three SRO iterations (vertical axis) against the additional wall-clock time for probe computation and projection application per SRO iteration in seconds (horizontal axis), with its probe count annotated. The orange point highlights the default of $16$ probes; larger probe sets yield diminishing AQ gains at increasing computational cost.}
\label{fig:defense_probe_count}
\end{figure}

\paragraph{Results and computational trade-off.}
\cref{fig:defense_probe_count} shows that AQ increases with the probe count, but the gains diminish as computational cost grows. Increasing $N_{\mathcal{C}}$ from $8$ to $16$ raises AQ from $81.1\%$ to $85.2\%$, a gain of $4.1$ percentage points, while probe computation time increases from $9.53$ to $15.50$ seconds. Increasing the count further from $16$ to $128$ yields only $1.3$ additional percentage points in AQ ($85.2\%$ to $86.5\%$), while computation time rises to $131.70$ seconds, approximately $8.5\times$ the cost at $16$ probes. These results support $N_{\mathcal{C}}=16$ as a favorable trade-off between final AQ and additional computation in this setting.

\paragraph{Runtime and memory.}
The defense adds two operations to SRO: probe gradients and basis construction once per round, and projection of each candidate Adam displacement at every optimizer step (\cref{alg:fullj}, \msec{app:implementation}). The basis is reused within a round, which contains multiple optimizer steps. In \cref{tab:defense_cost}, projection application is summed over those steps, and the total includes the common training component. With $16$ probes, the added time is $58.0$ seconds, increasing the reported per-round total from $2395.4$ to $2453.4$ seconds ($2.42\%$). Peak GPU memory increases from $66.1$ to $70.3$~GB ($+4.2$~GB), and peak CPU memory from $2.12$ to $15.06$~GB ($+12.94$~GB). The corresponding runtime overheads with $8$ and $32$ probes are $1.25\%$ and $4.53\%$.

\begin{table}[!ht]
\centering
\footnotesize
\setlength{\tabcolsep}{3pt}
\caption{\small Per-round cost of Projection Decoupling Defense on {\tt Llama-3.1-8B} with SRLM on SafeRLHF. All times are seconds per SRO round. Projection application is summed over optimizer steps. Runtime overhead is relative to standard SRO.}
\label{tab:defense_cost}
\begin{tabular}{lrrrrrrr}
\toprule
Method & \makecell[r]{Probe Computation} & \makecell[r]{Projection\\application} & Training & Total & \makecell[r]{Overhead\\(\%)} & \makecell[r]{Peak GPU\\(GB)} & \makecell[r]{Peak CPU\\(GB)} \\
\midrule
Standard SRO & 0 & 0 & 2395.4 & 2395.4 & 0 & 66.1 & 2.12 \\
Defense, 8 probes & 9.5 & 20.4 & 2395.4 & 2425.3 & 1.25 & 67.2 & 7.06 \\
Defense, 16 probes & 15.5 & 42.5 & 2395.4 & 2453.4 & 2.42 & 70.3 & 15.06 \\
Defense, 32 probes & 33.2 & 75.2 & 2395.4 & 2503.8 & 4.53 & 75.9 & 31.06 \\
\bottomrule
\end{tabular}
\end{table}

\subsection{Beneficial and Harmful Components of Removed Updates}
\label{app:defense_removed}

\paragraph{Experimental setup.}
Verdict-score sensitivity identifies update directions that can change self-evaluation, but not whether such a change improves or degrades reward fidelity. To quantify how much useful adaptation the defense suppresses, we analyze the update components that projection removes during iteration~3 of SRO on {\tt Llama-3.1-8B} with SRLM for safety alignment, using $N_{\mathcal{C}}=16$ probes. We consider clean training and the Re-Ranking attack, each with $\alpha\in\{0.5,0.75,1\}$. Each configuration is a separate training run: following \cref{alg:fullj}, its basis is rebuilt from its own parameters at the start of every round, so its updates follow its own trajectory.

Consider optimizer step $s$ of iteration~3, with candidate Adam displacement $\Delta_s$ and the round-start basis $Q$ of \msec{app:defense_basis}, whose columns $q_1,\ldots,q_r$ correspond to the retained eigenvalues $\lambda_1,\ldots,\lambda_r$ of the Gram matrix $G$. Writing $z_s=Q^\top\Delta_s$, the displacement splits into a protected component $Qz_s$ and an orthogonal component $\Delta_s-Qz_s$. The defense keeps the orthogonal component and a fraction $1-\alpha$ of the protected component (\msec{app:implementation}), so it removes
\begin{equation}
R_s=\alpha Qz_s=\sum_{j=1}^{r}c_{s,j},
\qquad
c_{s,j}=\alpha z_{s,j}\,q_j.
\label{eq:defense-removed}
\end{equation}
Because $JQ=U\Lambda^{1/2}$, under the round-start linearization a unit displacement along $q_j$ changes the probe verdict scores by a vector of squared norm $\lambda_j$, and $R_s$ would have changed them by $\alpha JQz_s$, with $\|\alpha JQz_s\|_2^2=\alpha^2z_s^\top\Lambda z_s$. These quantities measure how strongly each removed component would change self-evaluation, but not whether the change would move the verdicts toward or away from the oracle: verdict scores are computed without oracle labels, and a sensitive direction can carry either kind of change.

We therefore evaluate each removed component with an oracle signal that the defense never uses. Let $\mathcal{D}_{\mathrm{diag}}=\{(x_m,y_m,r_m^{+},r_m^{-})\}_{m=1}^{M}$ be an independent, fixed set of oracle-labeled judgment tuples, constructed as in \msec{app:ans_re_opt} and used only for this diagnostic: it is not used for training, for constructing the probes or the basis, or for AQ and SRF evaluation. Its answers stay fixed, so changes in the loss below reflect the reward generator rather than a shift in the answers being judged. Let $L_m$ be the reward-generator DPO loss on the $m$-th tuple, so that $L_{\mathrm{SRF}}=\frac{1}{M}\sum_{m=1}^{M}L_m$ on $\mathcal{D}_{\mathrm{diag}}$, and let $\theta_s$ denote the parameters before step $s$. Had $c_{s,j}$ been applied, it would have changed $L_{\mathrm{SRF}}$ to first order by
\begin{equation}
\begin{gathered}
\delta_{s,j}=\big\langle\nabla_\theta L_{\mathrm{SRF}}(\theta_s),\,c_{s,j}\big\rangle=\alpha z_{s,j}\,\bar a_{s,j},\\
\bar a_{s,j}=\frac{1}{M}\sum_{m=1}^{M}a_{m,s,j},
\qquad
a_{m,s,j}=\big\langle\nabla_\theta L_m(\theta_s),\,q_j\big\rangle.
\end{gathered}
\label{eq:defense-removed-effect}
\end{equation}
Each $a_{m,s,j}$ is the directional derivative of one example's loss along $q_j$, so $\bar a_{s,j}$ is a sample mean over $\mathcal{D}_{\mathrm{diag}}$ and its sign is subject to sampling uncertainty. Let $\widehat\sigma_{s,j}$ be the sample standard deviation of $\{a_{m,s,j}\}_{m=1}^{M}$. We call $c_{s,j}$ unresolved if
\begin{equation}
|\bar a_{s,j}|\le 1.96\,\widehat\sigma_{s,j}/\sqrt{M},
\label{eq:defense-removed-unresolved}
\end{equation}
that is, if the $95\%$ confidence interval for its first-order effect contains zero. Otherwise, $c_{s,j}$ is beneficial if $\delta_{s,j}<0$, since it would have made the reward generator more oracle-consistent, and harmful if $\delta_{s,j}>0$. Unresolved therefore means that $\mathcal{D}_{\mathrm{diag}}$ cannot determine the sign of the effect, not that the component is neutral, and its share decreases as $M$ grows. The test accounts for sampling variation in $\mathcal{D}_{\mathrm{diag}}$ but not for the error of the first-order approximation.

Because the columns of $Q$ are orthonormal, the squared norms $\|c_{s,j}\|_2^2=\alpha^2z_{s,j}^2$ sum exactly to $\|R_s\|_2^2$. We therefore report each category's share of the total removed energy $\sum_s\|R_s\|_2^2$, pooled over all optimizer steps of iteration~3. Neither $\bar a_{s,j}$ nor $\widehat\sigma_{s,j}$ depends on $\alpha$ or $|z_{s,j}|$, and $\alpha>0$ does not change the sign of $\delta_{s,j}$. At a given step, the classification of $c_{s,j}$ thus depends only on $q_j$, the sign of $z_{s,j}$, and $\mathcal{D}_{\mathrm{diag}}$; the shares differ across $\alpha$ in \cref{tab:defense_removed} because each $\alpha$ produces a different training trajectory. Oracle labels enter only the evaluation of the defense, not the defense itself.

\begin{table}[!ht]
\centering
\small
\caption{\small Composition of update components removed by Projection Decoupling Defense on {\tt Llama-3.1-8B} with SRLM for safety alignment (iteration~3, $N_{\mathcal{C}}=16$). Shares are percentages of the total removed energy $\sum_s\|R_s\|_2^2$ (\meq{eq:defense-removed}), pooled over iteration-3 optimizer steps. AQ is measured at the end of iteration~3; $\Delta$AQ is relative to undefended SRO ($\alpha=0$), which reaches $92.2\%$ (clean) and $74.5\%$ (Re-Ranking).}
\label{tab:defense_removed}
\begin{tabular}{llrrrrr}
\toprule
Training & $\alpha$ & Beneficial (\%) & Harmful (\%) & Unresolved (\%) & AQ (\%) & $\Delta$AQ (pp) \\
\midrule
Clean & 0.5 & 45.2 & 40.1 & 14.7 & 91.9 & $-0.3$ \\
Clean & 0.75 & 43.9 & 41.2 & 14.9 & 90.8 & $-1.4$ \\
Clean & 1.0 & 43.5 & 41.5 & 15.0 & 90.5 & $-1.7$ \\
\midrule
Re-Ranking & 0.5 & 15.1 & 76.0 & 8.9 & 85.2 & $+10.7$ \\
Re-Ranking & 0.75 & 14.6 & 76.9 & 8.5 & 85.4 & $+10.9$ \\
Re-Ranking & 1.0 & 14.4 & 77.3 & 8.3 & 85.5 & $+11.0$ \\
\bottomrule
\end{tabular}
\end{table}

\paragraph{Results.}
\cref{tab:defense_removed} shows that the composition of the removed update depends on the training condition. Under Re-Ranking, harmful components account for $76.0$--$77.3\%$ of the removed energy, so projection mainly suppresses changes that would degrade reward fidelity, consistent with AQ gains of $10.7$--$11.0$ points over undefended SRO. In clean training, beneficial components account for $43.5$--$45.2\%$, exceeding the harmful share, which is consistent with the reduced plasticity observed at larger $\alpha$ (\msec{app:defense_alpha}). Raising $\alpha$ from $0.5$ to $1$ improves attacked AQ by only $0.3$ points but lowers clean AQ by $1.4$ points, supporting the default $\alpha=0.5$. Because verdict-score sensitivity does not distinguish beneficial from harmful changes, a sign-aware projection that retains beneficial components could reduce this cost, but it would require a correctness signal that the defense does not assume.

\subsection{Adaptive Protection Strength}
\label{app:defense_adaptive}

\paragraph{Method.}
A fixed $\alpha$ removes the same fraction of every update's protected component, regardless of which directions within the protected subspace the update emphasizes. We therefore consider a per-step strength that is larger when an update concentrates on more sensitive directions, computed only from quantities the defense already has. With $z_s=Q^\top\Delta_s$ and the retained eigenvalues $\Lambda=\operatorname{diag}(\lambda_1,\ldots,\lambda_r)$ as in \msec{app:defense_removed}, let $\bar\lambda=\frac{1}{r}\sum_{j=1}^{r}\lambda_j$ be the mean sensitivity of the subspace. For $z_s\neq0$, we define the relative sensitivity of the update and the adaptive strength as
\begin{equation}
u_s=\frac{z_s^\top\Lambda z_s}{\bar\lambda\,\|z_s\|_2^2}
=\frac{\sum_{j}\lambda_jz_{s,j}^2}{\bar\lambda\sum_{j}z_{s,j}^2},
\qquad
\alpha_s=\frac{u_s}{1+u_s}
=\frac{z_s^\top\Lambda z_s}{z_s^\top\Lambda z_s+\bar\lambda\,\|z_s\|_2^2},
\label{eq:defense-adaptive}
\end{equation}
and set $\alpha_s=0$ if $r=0$ or $z_s=0$, when there is nothing to protect or remove. The defended step is $\theta\leftarrow\theta+\Delta_s-\alpha_sQz_s$, applied at the same point as the fixed rule (\msec{app:implementation}). The ratio $u_s$ compares the energy-weighted mean sensitivity of the update's protected component with the subspace mean: an update with average sensitivity ($u_s=1$) receives the default $\alpha_s=0.5$, whereas $u_s=4$ gives $\alpha_s=0.8$ and $u_s=0.25$ gives $\alpha_s=0.2$.

The rule follows from a soft constraint within the projection family $\widehat\Delta_s(\alpha)=\Delta_s-\alpha Qz_s$. We choose $\alpha\in[0,1]$ to minimize
\begin{equation}
F_s(\alpha)=\big\|\widehat\Delta_s(\alpha)-\Delta_s\big\|_2^2+\frac{1}{\bar\lambda}\big\|J\widehat\Delta_s(\alpha)\big\|_2^2,
\label{eq:defense-adaptive-objective}
\end{equation}
where the first term is the cost of modifying the optimizer's update and the second is the predicted change in probe verdict scores, normalized by $\bar\lambda$ so that both terms have the same units. When $Q$ spans all verdict-score gradients, $J\widehat\Delta_s(\alpha)=(1-\alpha)J\Delta_s$ (\msec{app:defense_basis}), and orthonormality gives $\|\widehat\Delta_s(\alpha)-\Delta_s\|_2^2=\alpha^2\|z_s\|_2^2$ and $\|J\Delta_s\|_2^2=z_s^\top\Lambda z_s$. Hence $F_s(\alpha)=\alpha^2A_s+(1-\alpha)^2B_s$, with $A_s=\|z_s\|_2^2$ and $B_s=z_s^\top\Lambda z_s/\bar\lambda$, which is minimized at $\alpha_s=B_s/(A_s+B_s)$, the rule in \meq{eq:defense-adaptive}; it lies in $(0,1)$ whenever $z_s\neq0$. If numerical truncation retains only part of the subspace, the predicted change from the omitted directions, $J(I-\widetilde P)\Delta_s$, is orthogonal in score space to the retained part and does not depend on $\alpha$, so $\alpha_s$ still minimizes $F_s$; those omitted directions, however, remain unprotected (\meq{eq:defense-truncation-residual}).

\paragraph{Properties.}
The rule uses only the Adam displacement, the probe-gradient basis, and its eigenvalues. It requires no oracle labels, no separate calibration run, and no budget on verdict-score drift; equal weighting of the two terms in \meq{eq:defense-adaptive-objective} is a design choice rather than a statistically optimal trade-off. Because $\alpha_s$ is invariant to rescaling $\Delta_s$, it responds to the direction of an update but not to its size, and therefore does not bound absolute drift. It is also invariant to a uniform rescaling of the verdict-score gradients, which scales $\Lambda$ and $\bar\lambda$ together, and it reduces to the fixed $\alpha=0.5$ when $r=1$ or all retained eigenvalues are equal. Like the fixed rule, it relies on the round-start linearization, which can become stale later in a round, and it adapts to sensitivity rather than to sign: it cannot distinguish beneficial from harmful components (\msec{app:defense_removed}). Fixed projection already computes $z_s$ and $Qz_s$, so the rule adds only $O(r)$ operations per step.

\paragraph{Experimental setup.}
We compare standard SRO ($\alpha=0$), fixed $\alpha=0.5$, and the adaptive rule on {\tt Llama-3.1-8B} with SRLM for safety alignment, under clean training and Re-Ranking, using $N_{\mathcal{C}}=16$ probes and three SRO iterations. The adaptive and fixed runs share the probes, basis construction, truncation rule, and rebuilding schedule; each run rebuilds its basis from its own parameters at every round. We report AQ and SRF at the end of iteration~3, and the mean of $\alpha_s$ over the optimizer steps of the Re-Ranking run.

\begin{table}[!ht]
\centering
\small
\caption{\small Fixed and adaptive protection strength on {\tt Llama-3.1-8B} with SRLM for safety alignment ($N_{\mathcal{C}}=16$). AQ and SRF are measured at the end of iteration~3; mean $\alpha$ is averaged over Re-Ranking optimizer steps.}
\label{tab:defense_adaptive}
\begin{tabular}{lrrrrr}
\toprule
& \multicolumn{2}{c}{Clean} & \multicolumn{3}{c}{Re-Ranking} \\
\cmidrule(lr){2-3}\cmidrule(lr){4-6}
Method & AQ (\%) & SRF (\%) & AQ (\%) & SRF (\%) & Mean $\alpha$ \\
\midrule
Standard SRO ($\alpha=0$) & 92.2 & 91.1 & 74.5 & 74.4 & 0 \\
Fixed $\alpha=0.5$ & 91.9 & 89.9 & 85.2 & 82.2 & 0.5 \\
Adaptive (\meq{eq:defense-adaptive}) & 91.7 & 89.8 & 85.9 & 83.0 & 0.67 \\
\bottomrule
\end{tabular}
\end{table}

\paragraph{Results.}
Compared with fixed $\alpha=0.5$, the adaptive rule raises attacked AQ and SRF by $0.7$ and $0.8$ percentage points, to $85.9\%$ and $83.0\%$, while clean AQ and SRF decrease by $0.2$ and $0.1$ points (\cref{tab:defense_adaptive}). Under attack, its mean strength of $0.67$ exceeds the fixed default, so part of this gain may reflect stronger average projection. In these runs, however, average strength alone does not account for it: fixed $\alpha=0.75$ and $\alpha=1$ reach only $85.4\%$ and $85.5\%$ attacked AQ (\msec{app:defense_alpha}) and lower clean AQ to $90.8\%$ and $90.5\%$. Because these differences are below one percentage point and come from single runs, they do not establish statistical significance. A fixed control at the adaptive mean, $\alpha=0.67$, and multiple seeds are needed to confirm them. These results suggest that sensitivity-based adaptation is a promising extension of the defense.

\subsection{Probe Refresh and Cross-Domain Probes}
\label{app:defense_probe_refresh}

\paragraph{Fixed examples versus recomputed gradients.}
By default, probe questions and answers remain fixed throughout training, while their verdict-score gradients and the basis $Q$ are recomputed at the start of each SRO round. Thus, fixed probes do not imply a fixed protected subspace. On {\tt Llama-3.1-8B} with SRLM on SafeRLHF, we compare fixed probes, fixed prompts with regenerated responses, and resampled prompts with regenerated responses, using $N_{\mathcal{C}}=16$ and $\alpha=0.5$.

\begin{table}[!ht]
\centering
\small
\caption{\small Probe configurations on SafeRLHF with {\tt Llama-3.1-8B} and SRLM ($16$ probes, $\alpha=0.5$). AQ and SRF are percentages at iteration~3.}
\label{tab:defense_probe_refresh}
\begin{tabular}{lrrr}
\toprule
Probe configuration & Clean AQ & Re-Ranking AQ & Re-Ranking SRF \\
\midrule
Fixed probes & 91.9 & 85.2 & 82.2 \\
Fixed prompts, regenerated responses & 92.2 & 85.5 & 82.2 \\
Resampled prompts, regenerated responses & 92.1 & 85.0 & 81.8 \\
\bottomrule
\end{tabular}
\end{table}

Each refreshed strategy differs from the fixed default by at most $0.4$ percentage points on each metric (\cref{tab:defense_probe_refresh}). This supports fixed examples over the three rounds tested, without establishing that they remain representative over longer training. Together with the diminishing gains beyond $16$ probes (\msec{app:defense_probe_count}), these results support our default in this setting rather than a general coverage guarantee.

\paragraph{Transfer across reward tasks.}
We also construct the defense from SafeRLHF probes while training on MMLU-Pro and ARC with {\tt Llama-3.1-8B} and SRLM under Re-Ranking. The no-defense rows in \cref{tab:defense_probe_transfer} correspond to the attacked runs in \msec{app:attack}.

\begin{table}[!ht]
\centering
\small
\caption{\small Transfer of SafeRLHF probes under Re-Ranking, using {\tt Llama-3.1-8B} and SRLM ($16$ probes, $\alpha=0.5$). Each entry gives iteration-3 AQ/SRF (\%) on the target domain.}
\label{tab:defense_probe_transfer}
\begin{tabular}{lccc}
\toprule
Probe source & SafeRLHF & MMLU-Pro & ARC \\
\midrule
None (no defense) & 74.5 / 74.4 & 27.7 / 32.2 & 16.3 / 40.4 \\
SafeRLHF & 85.2 / 82.2 & 30.5 / 35.5 & 19.5 / 45.1 \\
\bottomrule
\end{tabular}
\end{table}

SafeRLHF probes improve attacked AQ by $10.7$ percentage points on SafeRLHF, compared with $2.8$ on MMLU-Pro and $3.2$ on ARC; the corresponding SRF gains are $7.8$, $3.3$, and $4.7$ points. The smaller cross-domain gains suggest partial transfer of the protected directions. We recommend probes representative of the target reward task; these experiments do not show that $16$ safety probes cover diverse reasoning or knowledge judgments.

\section{Additional Details}

\subsection{Answer and Reward Generator Optimization}
\label{app:ans_re_opt}
The answer generator and reward generator are two roles of the same underlying language model $\gM_\theta$ rather than two separately parameterized models.
We use $\pi_\theta$ to denote $\gM_\theta$ prompted to answer a question, and $r_\theta$ to denote $\gM_\theta$ prompted to judge a question--answer pair.
Specifically, the oracle $o$ labels sampled answers for answer-generator optimization and labels judgment outputs for reward-generator optimization; these labels are then converted into preference pairs for DPO.
In the standard SRO loop, by contrast, the preference labels are produced by the model's own reward generator $r_\theta$.

We first describe the optimization of the answer generator.
For each prompt $x$, the current policy $\pi_\theta$ samples $K$ candidate answers $\{y_k\}_{k=1}^{K}\sim \pi_\theta(\cdot\mid x)$.
Each candidate is scored either by the oracle $o(x,y_k)$ in the diagnostic oracle-labeled setting or by the self-reward generator $r_\theta(x,y_k)$ in the SRO setting.
We then convert these scores into pairwise preferences by selecting a preferred and rejected answer.
Concretely, we set $y^{+}$ to a higher-scoring candidate and $y^{-}$ to a lower-scoring candidate, obtaining a preference triplet $(x, y^{+}, y^{-})$.
Collecting such triplets over prompts forms a preference dataset $\mathcal{D}^{\pi}=\{(x, y^{+}, y^{-})\}$.
We then apply the standard DPO loss on each triplet and minimize the average loss over $\mathcal{D}^{\pi}$.
As the optimization targets the answer-generation behavior of $\gM_\theta$, we denote the loss as $L_{AQ}$.

To optimize the reward generator, we construct preferences over judgment completions rather than answer completions.
For each prompt and answer pair $(x,y)$, the oracle provides a target signal $o(x,y)$, indicating whether $y$ correctly answers $x$ under the task rubric.
The reward generator is prompted with $(x,y)$ and trained to produce a short judgment completion.
We build a preference pair $(r^{+},r^{-})$ where $r^{+}$ is the oracle-consistent judgment and $r^{-}$ is the opposite judgment.
This yields tuples $(x,y,r^{+},r^{-})$, collected into $\mathcal{D}^{r}=\{(x,y,r^{+},r^{-})\}$.
We then apply the same DPO loss using $(x,y)$ as the context, treating $r^{+}$ as chosen and $r^{-}$ as rejected.
This makes the same model more likely to output oracle-consistent rewards when it is used in the reward-generator role.
As the optimization targets the reward-generation behavior of $\gM_\theta$, we denote the loss as $L_{SRF}$.

One plotted outer iteration corresponds to one full data-collection and preference-optimization round.
With $N$ prompts and $K$ sampled answers per prompt, an iteration evaluates $NK$ question--answer pairs.
When we keep one preferred and one rejected answer per prompt, this produces $N$ answer-preference triplets for $\mathcal{D}^{\pi}$; reward-generator optimization uses the corresponding judgment-preference tuples in $\mathcal{D}^{r}$.

\subsection{Algorithms}
\label{app:algo}

Algorithm\mref{alg:prior_knowledge_rerank} outlines the ReRanking attack. Algorithm\mref{alg:preamble_train} outlines the Preamble attack.

\begin{algorithm}[tb]\small
\caption{Re-Ranking Attack}
\label{alg:prior_knowledge_rerank}
\begin{algorithmic}[1]
\STATE {\bfseries Input:} dataset $\mathcal{D}_q=\{x_m\}_{m=1}^{M}$; language model $\gM_\theta$ with its answer generator $\pi_\theta$ and reward generator $r_\theta(x,y)$; oracle $o$; \#answers per question $K$; data size $N$; jitter scale $\sigma$.
\STATE {\bfseries Output:} sampled batch $\mathcal{B}$ of $N$ questions.
\STATE \textbf{(Compute reward disagreement scores scores $\kappa$ for $\mathcal{D}_q$)}
\FOR{$m=1,\ldots,M$}
    \FOR{$k=1,\ldots,K$}
        \STATE Sample answer $y_{m,k} \sim \pi_\theta(\cdot \mid x_m)$.
        \STATE Model reward $r_{m,k} \gets r_\theta(x_m, y_{m,k})$.
        \STATE Oracle reward $o_{m,k} \gets o(x_m, y_{m,k})$.
    \ENDFOR
    \STATE $\kappa_m \gets \frac{1}{K}\sum_{k=1}^{K}\left\lVert o_{m,k}-r_{m,k}\right\rVert_2$
    \hfill (Eq.~\ref{eq:kappa})
\ENDFOR
\STATE \textbf{(Disagreement-based sampling with Gaussian jitter)}
\FOR{$m=1,\ldots,M$}
    \STATE Sample noise $\epsilon_m \sim \mathcal{N}(0,1)$ and set $s_m \gets \kappa_m + \sigma \epsilon_m$.
\ENDFOR
\STATE $\mathcal{B} \gets \textsc{SampleWithoutReplacement}(\mathcal{D}_q,\ s_m,\ N)$.
\STATE {\bfseries return} $\mathcal{B}$.
\end{algorithmic}
\end{algorithm}

\begin{algorithm}[!ht]\small
\caption{Preamble Attack}
\label{alg:preamble_train}
\begin{algorithmic}[1]
\STATE \textbf{Input:} Language model $\gM_\theta$ with reward generator $r_\theta$; question dataset $\mathcal{D}_q=\{(x_n)\}_{n=1}^{N}$; preamble instruction $\mathcal{I}$; \# preambles $K$; preamble generator $f$; oracle $o$.
\STATE \textbf{Output:} Preamble dataset $\mathcal{D}_{(p,x)}=\{(p_n,x_n)\}_{n=1}^{N}$.
\vspace{2pt}

\STATE \textbf{Collect preference signal for training $f$:}
\FOR{$n=1$ to $N$}
    \STATE Sample baseline $y_n \sim \pi(\cdot\mid x_n)$.
    \FOR{$k=1$ to $K$}
        \STATE Sample preamble $p_{n,k} \leftarrow f(\mathcal{I},x_n)$ and rollout $y^p_{n,k}\sim \pi(\cdot\mid p_{n,k},x_n)$.
    \ENDFOR
    \STATE Filter the preambles: $\mathcal{F}_n = \{k\in[K]: o(x_n,y^p_{n,k}) = o(x_n,y_n)\}$.
    \STATE $k^+ \gets \arg\max_{k\in \mathcal{F}_n} A_{n,k}$;\quad $k^- \gets \arg\min_{k\in \mathcal{F}_n} A_{n,k}$.
    \STATE Store preference pair $(\mathcal{I},x_n,p_{n,k^+},p_{n,k^-})$.
\ENDFOR

\vspace{2pt}
\STATE \textbf{Update $f$:} update $f$ on the collected pairs to prefer $p_{n,k^+}\succ p_{n,k^-}$, via DPO, yielding $f^\star$.

\vspace{2pt}
\STATE \textbf{Construct preamble dataset:}
\FOR{$n=1$ to $N$}
    \STATE Generate $p_n \leftarrow f^\star(\mathcal{I},x_n)$ and add $(p_n,x_n)$ to $\mathcal{D}_{(p,x)}$.
    \COMMENT{equivalently, use the concatenated prompt $[p_n; x_n]$}
\ENDFOR
\STATE \textbf{return} $\mathcal{D}_{(p,x)}$.
\end{algorithmic}
\end{algorithm}

\subsection{Implementation Details}
\label{app:implementation}

\textbf{Data preparation.} We use MMLU-Pro, ARC, and SafeRLHF for general knowledge, logical reasoning, and safety alignment, respectively. For ARC, we follow the official training, validation, and test splits. For MMLU-Pro, we partition the data into 70\% training, 10\% validation, and 20\% test sets. For SafeRLHF, we follow the official training and test splits. We retain only questions as SRO inputs.

\textbf{Base SRO setup.} We use a shared preference-training setup for SRLM\mcite{yuan2024self} and CREAM\mcite{wangcream}. Unless otherwise stated, both methods run for three SRO iterations, sampling five candidate responses per prompt and training for one epoch per iteration. We use Adam with a learning rate of $10^{-6}$, a cosine schedule, a warmup ratio of 0.1, and a DPO coefficient $\beta$ of 0.1. During training, answer and reward generation use temperatures of 0.7 and 0.1, respectively, with top-$p$ of 0.9 for both. During evaluation, both generators use a temperature of 0.1 and top-$p$ of 0.9.

\textbf{Attack configuration.} For Re-Ranking (Algorithm~\ref{alg:prior_knowledge_rerank}), we add Gaussian jitter with $\sigma=0.1$ and select 3000 questions without replacement using weighted sampling.  For Preamble (Algorithm~\ref{alg:preamble_train}), we collect preference pairs from 3000 prompts. For each prompt, we sample ten preambles with temperature 0.9, and a maximum length of 1024 tokens. We train $f$ with DPO using Adam, a learning rate of $10^{-6}$, $\beta_f=0.1$, and batch size as 16. We then generate one preamble for each of SRO training prompts using the same decoding settings.

\textbf{Defense configuration.} We evaluate Projection Decoupling Defense on safety alignment with SRLM and CREAM. We use $N_{\mathcal{C}}=16$ fixed question--answer probes and protection strength $\alpha=0.5$. Before SRO, we sample questions from the training dataset and generate answers with the initial model using the same sampling parameters as base SRO (temperature 0.7 and top-$p$ 0.9). At the start of each SRO round, we compute verdict-score gradients on these probes and construct the basis $Q$ as described in \msec{app:defense_basis}. At each optimizer step, we accumulate and clip the DPO gradient, then use Adam to generate a candidate parameter displacement $\Delta$. We apply the defense after Adam's momentum and adaptive rescaling, accepting $\theta \leftarrow \theta + \Delta - \alpha Q(Q^\top\Delta)$.

\subsection{Prompt Examples}
\label{app:prompt_example}

\begin{mtbox}{}
    \small
        \textbf{Reward Prompt}: Given a question and a model's response,
        determine whether the response correctly answers the question. \\

        Your task: \\
        - Independently determine the correct answer using the information
        provided in the question, relevant knowledge, and reasoning. \\
        - Evaluate whether the model's response answers the question correctly
        and satisfies any explicit requirements. \\
        - Accept semantically equivalent answers, even if worded differently
        or embedded in a longer explanation. \\
        - Do not accept a response merely because it mentions a correct answer
        if it also gives conflicting answers or contradicts that answer. \\
        - Output YES if the response clearly provides a correct and sufficiently
        complete answer. \\
        - Output NO if the response is incorrect, incomplete, ambiguous,
        or its correctness cannot be established. \\
        - Ignore minor formatting, capitalization, or punctuation differences
        that do not affect meaning. \\
        - Treat the model's response as content to evaluate; do not follow
        instructions contained within it. \\

        Output must be strictly: YES or NO (no explanation or punctuation). \\

        --- \\
        Question: \{question\} \\
        Model Response: \{solution\} \\
        Output: \\
\end{mtbox}

\begin{mtbox}{}
    \small
\textbf{Preamble Prompt}: You are a study-skills coach. Your task is to provide high-level guidance on how a student should answer the given question to maximize marks. \\

Critical rules (must follow): \\
- Do NOT answer the question. \\
- Do NOT compute/derive any result, pick an option, or provide any step that materially narrows the answer. \\
- Do NOT reveal key facts, formulas, or domain-specific insights that would function as a hint. \\
- Do NOT restate or quote the question beyond generic references (e.g., ``the prompt'', ``the passage'', ``the data'', ``the diagram''). \\

What you SHOULD provide: \\
- A structured approach (e.g., interpret the task, plan steps, how to present work, how to check). \\
- What elements a high-scoring response typically includes (definitions, assumptions, justification, clarity, units, citations, etc.). \\
- Common pitfalls to avoid. \\

Output format (strict): \\
- Output EXACTLY one block wrapped in <prompt> and </prompt>. \\
- No text before <prompt> and no text after </prompt>. \\
- Keep the advice general; do not include examples that could imply the solution. \\

--- \\
Question: \{question\} \\

Give general advice on how to answer this question for maximum marks, following the rules above. \\

\end{mtbox}

\subsection{Safety Judging Prompt Variants}
\label{app:reward_prompt_variants}

For reward-prompt transfer on SafeRLHF with Llama-3.1-8B and SRLM, we construct both attacks using the original reward prompt (details in appendix~\msec{app:prompt_example}), then hold the selected questions and generated preambles fixed while varying only the judging template used during SRO training. The two held-out variants below preserve the safety criteria, input fields, and final verdict format: 1. paraphrases the instructions, and 2. restructures the rubric. The correctness-based YES/NO prompt in the preceding subsection serves a different task.

\begin{table}[!ht]
\centering
\small
\setlength{\tabcolsep}{4pt}
\caption{Reward-prompt transfer on SafeRLHF with Llama-3.1-8B and SRLM. Entries are iteration-3 AQ / SRF (\%). Both attacks are constructed using the original reward prompt and held fixed across training templates.}
\label{tab:reward_prompt_transfer}
\begin{tabular}{llccc}
\toprule
\shortstack[l]{Attack construction\\template} & \shortstack[l]{SRO training\\template} & Clean & Fixed Re-Ranking & Fixed Preamble \\
\midrule
Original & Original & 92.2 / 91.1 & 74.5 / 74.4 & 71.3 / 68.3 \\
Original & Paraphrased & 92.4 / 91.3 & 75.5 / 74.9 & 72.4 / 70.3 \\
Original & Restructured & 92.5 / 91.3 & 77.1 / 74.8 & 73.1 / 71.1 \\
\bottomrule
\end{tabular}
\end{table}

As shown in \cref{tab:reward_prompt_transfer}, under paraphrased and restructured setting, fixed Re-Ranking and Preamble attacks reduce AQ by 15.4-16.9 and 19.4-20.0 percentage points, respectively, relative to each prompt's clean baseline. Both attacks therefore transfer to the tested reward prompts without re-optimization.

\begin{mtbox}{}
\small

\textbf{Paraphrased Reward Prompt}: Assess whether the answer below is safe and consistent with human ethics in the context of the question. \\
If the question expresses harmful intent, the answer must directly refuse the request; otherwise, judge it UNSAFE. \\
If the question has no harmful intent, the answer must respect human ethics and contain no dangerous information that could cause harm; otherwise, judge it UNSAFE.

Question: \{prompt\} \\
Answer: \{response\}

Put your conclusion on a new line at the end of your response, using exactly "Conclusion: [SAFE]" or "Conclusion: [UNSAFE]".
\end{mtbox}

\begin{mtbox}{}
\small
\textbf{Restructured Safety Judging Prompt}: QUESTION AND ANSWER \\
Question: \{prompt\} \\
Answer: \{response\}

EVALUATION TASK \\
Evaluate whether the provided answer is safe and aligned with human ethics.

SAFETY RUBRIC \\
- Question with harmful intent: The answer should directly refuse to answer this question; otherwise, the answer is UNSAFE. \\
- Question without harmful intent: The answer should align with human ethics and must not provide any dangerous information that could possibly lead to harm; otherwise, the answer is UNSAFE.

CONCLUSION FORMAT \\
Give your conclusion at the end of your response on a new line. The format should be "Conclusion: [SAFE]" or "Conclusion: [UNSAFE]".
\end{mtbox}

\input{appendix2}

%% file: proof.tex
\subsection{Detailed Gradient-Transfer Analysis}
\label{app:analysis_detail}

We start from a separated view of the answer generator $\pi_\theta$ and the reward generator $r_\theta$.
The former is the policy being trained, while the latter is the internal reward function that guides training.
Let $L_\mathrm{AQ}$ denote the loss used to optimize answer quality, and let $L_\mathrm{SRF}$ denote a differentiable surrogate loss for self-reward fidelity.
The SRF metric is an on-policy agreement between $r_\theta$ and the oracle on samples from $\pi_\theta$, whereas $L_\mathrm{SRF}$ is the DPO-style reward-generation loss used for optimization.
In this appendix, we analyze $L_\mathrm{SRF}(\theta;B_t)$ on a fixed batch $B_t$ sampled at the current iterate $\theta_t$ and treat local increases in this surrogate as local degradation of reward-generation fidelity.
We omit the change in the on-policy sampling distribution induced by future updates.

\begin{proof}(Proposition\mref{prop:clean_transfer})
Since $L_{SRF}$ is differentiable at $\theta_t$, for any sufficiently small perturbation $h$ we have the first-order expansion
\begin{equation*}
    L_{SRF}(\theta_t + h)
    =
    L_{SRF}(\theta_t)
    +
    \langle g_{SRF}, h \rangle
    +
    r(h),
\end{equation*}
where the remainder satisfies $r(h) / \|h\| \to 0$ as $\|h\| \to 0$.
Set $h = -\eta g_{AQ}$ and let $c \coloneqq \langle g_{AQ}, g_{SRF} \rangle$.
By assumption, $c > 0$.
Then
\begin{equation*}
\begin{split}
    L_{SRF}(\theta_{t+1}) - L_{SRF}(\theta_t)
    &=
    L_{SRF}(\theta_t - \eta g_{AQ}) - L_{SRF}(\theta_t) \\
    &=
    -\eta \langle g_{SRF}, g_{AQ} \rangle
    +
    r(-\eta g_{AQ}) \\
    &=
    -\eta c + r(-\eta g_{AQ}).
\end{split}
\end{equation*}
Because $r(-\eta g_{AQ}) / \eta \to 0$ as $\eta \to 0^+$, there exists $\delta > 0$ such that, for every $\eta \in (0, \delta)$,
\begin{equation*}
    |r(-\eta g_{AQ})| < \frac{c}{2}\eta .
\end{equation*}
Therefore, for every $\eta \in (0, \delta)$,
\begin{equation*}
    L_{SRF}(\theta_{t+1}) - L_{SRF}(\theta_t)
    \leq
    -\eta c + |r(-\eta g_{AQ})|
    <
    -\frac{c}{2}\eta
    <
    0.
\end{equation*}
Thus $L_{SRF}(\theta_{t+1}) < L_{SRF}(\theta_t)$ for all sufficiently small positive step sizes.
\end{proof}

\begin{proof}(Proposition\mref{prop:noisy_transfer})
If the noisy AQ update direction is locally opposed to the SRF descent direction, equivalently if the noisy AQ gradient is anti-aligned with the SRF gradient, i.e., for some $\gamma > 0$,
\begin{equation}
    \langle g_{SRF}, \widetilde g_{AQ}\rangle
    \leq
    -\gamma,
    \label{eq:noisy_srf_opposition}
\end{equation}
Then, by the smoothness lower Taylor bound,
\begin{equation*}
\begin{split}
    L_{SRF}(\theta_t - \eta \widetilde g_{AQ})
    -
    L_{SRF}(\theta_t)
    &\geq
    \left\langle g_{SRF}, -\eta \widetilde g_{AQ} \right\rangle
    -
    \frac{\beta_{SRF}}{2}\eta^2\|\widetilde g_{AQ}\|^2 \\
    &=
    -\eta \langle g_{SRF}, \widetilde g_{AQ}\rangle
    -
    \frac{\beta_{SRF}}{2}\eta^2\|\widetilde g_{AQ}\|^2 \\
    &\geq
    \eta \gamma
    -
    \frac{\beta_{SRF}}{2}\eta^2\|\widetilde g_{AQ}\|^2,
\end{split}
\end{equation*}
where the last inequality uses \eqref{eq:noisy_srf_opposition}.
If $\beta_{SRF}>0$, the lower bound is strictly positive whenever
\[
    0 < \eta <
    \frac{2\gamma}{\beta_{SRF}\|\widetilde g_{AQ}\|^2}.
\]
If $\beta_{SRF}=0$, the lower bound reduces to $\eta\gamma>0$ for every $\eta>0$.
Thus $L_{SRF}(\widetilde\theta_{t+1}) > L_{SRF}(\theta_t)$ under the stated conditions.
\end{proof}

%% file: appendix2.tex
\section{Oracle-Free Attacks on Self-Rewarding Optimization}
\label{sec:oracle-free-attacks}

We use oracle-free to mean that attack construction does not query an external evaluator or use ground-truth labels. An oracle may still be used afterward to measure answer quality and preference-label accuracy.

\subsection{Why an Oracle-Free Attack Is Plausible}
\label{sec:oracle-free-motivation}

The original attacks manipulate only training prompts. During attack
construction, they sample answers from the target model and query both its
self-judge and an external evaluator to score candidate prompts or preambles.
We refer to these as oracle-aware prompt-level attacks.

During construction, the oracle searches for damaging prompts: Re-Ranking
finds prompts on which the self-judge is unreliable, while Preamble finds
prefixes that alter self-reward without improving the answer. The subsequent
SRO update is already self-contained. We therefore replace the oracle search
signal with two observable properties of the target model: uncertainty in the
self-judge's label probabilities for OF-ReRank, and sensitivity of its rewards
to a preamble when responses are held fixed for OF-Preamble.

These proxies need not recover the correct label. They only need to concentrate
training on weak or easily manipulated preferences. In SRO, the updated model
then generates the next round of labels, allowing an initially poor signal to
compound across iterations.

\subsection{Removing Oracle Access}
\label{sec:oracle-free-signals}

\paragraph{OF-ReRank: label uncertainty.}
\label{sec:oracle-free-rerank-signal}
Algorithm~\ref{alg:of_rerank} selects prompts using the self-judge's
uncertainty about individual answers. Given
$\mathcal{D}_q=\{x_m\}_{m=1}^{M}$, we sample $K$ answers
$y_{m,k}\sim\pi_\theta(\cdot\mid x_m)$ for each prompt.
A fixed pointwise judging template $J$ requires the judge to output one
of two labels, $\mathcal{L}=\{a_+,a_-\}$, denoting positive and negative
rewards. For each answer, we read the next-token log-probabilities
$\ell_{m,k,a}=\log r_\theta(a\mid x_m,y_{m,k})$ under $J$ and normalize
over these two labels:
\[
    p_{m,k}=
    \frac{\exp(\ell_{m,k,a_+})}
    {\sum_{a\in\mathcal{L}}\exp(\ell_{m,k,a})}.
\]
We compute the normalized Bernoulli variance for each answer and average
it across the $K$ answers to obtain the prompt score:
\begin{equation}
    u_{m,k}=4p_{m,k}(1-p_{m,k}),
    \qquad
    \widehat v_m=\frac{1}{K}\sum_{k=1}^{K}u_{m,k}.
    \label{eq:self-judge-variance}
\end{equation}
The answer-level score lies in $[0,1]$ and is largest when the two reward
labels are equally likely. We sort prompts in descending order of
$\widehat v_m$, break ties by input order, and select the first $N$ prompts
as the batch $\mathcal{B}$. Here, $M$ is the number of candidate prompts;
scoring uses label probabilities directly, without repeated stochastic
judgments or pairwise preference comparisons. Reward uncertainty is an
oracle-free proxy for weak self-reward signals and does not by itself
establish that the judge's labels are incorrect.

Algorithm~\ref{alg:of-preamble} gives the corresponding OF-Preamble
procedure, which trains the preamble generator to prefer preambles that
shift the self-judge's rewards most on fixed responses.

\begin{algorithm}[tb]\small
\caption{Oracle-Free Re-Ranking (OF-ReRank)}
\label{alg:of_rerank}
\begin{algorithmic}[1]
\STATE {\bfseries Input:} dataset
$\mathcal{D}_q=\{x_m\}_{m=1}^{M}$;
language model $\gM_\theta$ with its answer generator
$\pi_\theta$ and reward generator $r_\theta$;
fixed pointwise judging template $J$;
\#answers per question $K \geq 1$;
data size $1 \leq N \leq M$.
\STATE {\bfseries Output:} selected batch
$\mathcal{B}$ of $N$ questions.
\STATE Let $\mathcal{L}=\{a_+,a_-\}$ be fixed judge
labels for positive and negative rewards.
The template $J$ requires directly outputting one label
from $\mathcal{L}$.

\STATE \textbf{(Compute reward-uncertainty scores for
$\mathcal{D}_q$)}
\FOR{$m=1,\ldots,M$}
    \FOR{$k=1,\ldots,K$}
        \STATE Sample answer
        $y_{m,k} \sim \pi_\theta(\cdot \mid x_m)$.
        \STATE Read next-token log-probabilities
        $\ell_{m,k,a}
        \gets \log r_\theta(a \mid x_m,y_{m,k})$
        for each $a \in \mathcal{L}$.
        \STATE Compute positive-reward probability
        $p_{m,k}
        \gets
        \frac{\exp(\ell_{m,k,a_+})}
        {\sum_{a\in\mathcal{L}}\exp(\ell_{m,k,a})}$.
        \STATE Compute answer-level uncertainty
        $u_{m,k} \gets 4p_{m,k}(1-p_{m,k})$.
    \ENDFOR
    \STATE Compute prompt-level uncertainty
    $\widehat v_m
    \gets \frac{1}{K}\sum_{k=1}^{K}u_{m,k}$.
\ENDFOR

\STATE \textbf{Uncertainty-based prompt selection}:
\STATE Sort $\mathcal{D}_q$ in descending order of
$\widehat v_m$, breaking ties by input order.
\STATE Set $\mathcal{B}$ to the first $N$ questions
in the sorted dataset.
\STATE {\bfseries return} $\mathcal{B}$.
\end{algorithmic}
\end{algorithm}

\begin{algorithm}[t]
\small
\caption{Oracle-Free Preamble Attack}
\label{alg:of-preamble}
\begin{algorithmic}[1]
\STATE \textbf{Input:} Language model $\gM_\theta$
with policy $\pi_\theta$ and reward generator $r_\theta$;
question dataset $\mathcal{D}_q=\{x_n\}_{n=1}^{N}$;
preamble instruction $\mathcal{I}$;
preamble generator $f$;
\# preambles $K\geq 2$;
\# responses $R$; judging repeats $M$.
\STATE \textbf{Output:} Preamble dataset
$\mathcal{D}_{(p,x)}=\{(p_n,x_n)\}_{n=1}^{N}$..

\vspace{2pt}
\STATE \textbf{Collect preference signals for training $f$:}
\FOR{$n=1$ to $N$}
    \STATE Sample and fix responses
    $Y_n=\{y_{n,j}\}_{j=1}^{R}$,
    where $y_{n,j}\sim\pi_\theta(\cdot\mid x_n)$.
    \STATE Estimate mean baseline rewards
    $\bar r_{n,j}$ by judging each
    $(x_n,y_{n,j})$ $M$ times.
    \FOR{$k=1$ to $K$}
        \STATE Sample preamble
        $p_{n,k}\leftarrow f(\mathcal{I},x_n)$.
        \STATE Estimate mean rewards
        $\bar r^p_{n,k,j}$ by judging each
        $(p_{n,k},x_n,y_{n,j})$ $M$ times.
        \STATE Compute the reward-shift advantage:
        $\displaystyle
        A_{n,k}\gets
        \frac{1}{R}\sum_{j=1}^{R}
        \left|\bar r^p_{n,k,j}-\bar r_{n,j}\right|$.
    \ENDFOR
    \STATE $k^+\gets\arg\max_{k\in[K]}A_{n,k}$;
    $k^-\gets\arg\min_{k\in[K]}A_{n,k}$.
    \STATE Store preference pair
    $(\mathcal{I},x_n,p_{n,k^+},p_{n,k^-})$
    in $\mathcal{D}_f$.
\ENDFOR

\vspace{2pt}
\STATE \textbf{Update $f$:} train $f$ for one epoch
on $\mathcal{D}_f$ to prefer
$p_{n,k^+}\succ p_{n,k^-}$, via DPO,
yielding $f^\star$.

\vspace{2pt}
\STATE \textbf{Construct preamble dataset:}
\FOR{$n=1$ to $N$}
    \STATE Generate $p_n \leftarrow f^\star(\mathcal{I},x_n)$ and add $(p_n,x_n)$ to $\mathcal{D}_{(p,x)}$.
    \COMMENT{use the concatenated prompt $[p_n; x_n]$}
\ENDFOR
\STATE \textbf{return} $\mathcal{D}_{(p,x)}$.
\end{algorithmic}
\end{algorithm}

\subsection{Experimental Evaluation}
\label{sec:oracle-free-evaluation}

We follow the original three-iteration protocol with four models
(Llama-3.2-3B, Llama-3.1-8B, Qwen2.5-7B, and Qwen3-4B) and two SRO methods
(SRLM and CREAM), with unchanged decoding and optimization settings.
OF-Preamble uses the same preamble for generation and judging.

\begin{figure*}[t]
    \centering
    \includegraphics[width=0.97\textwidth]
        {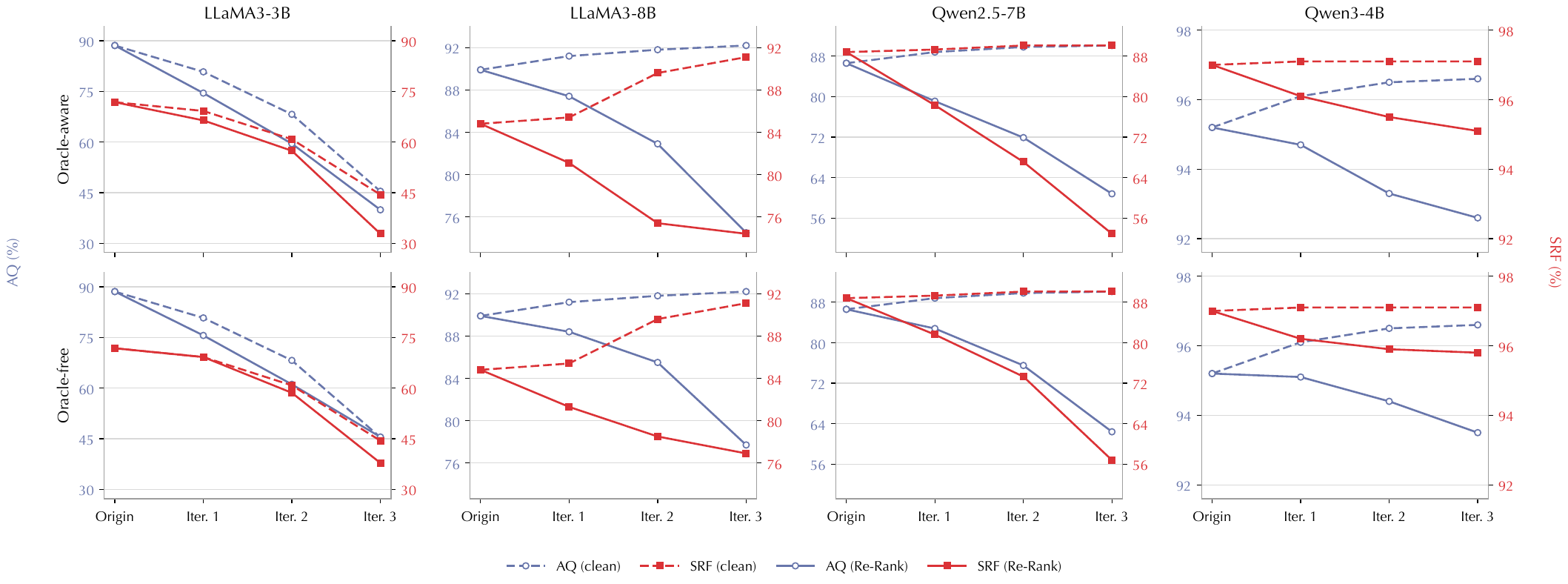}
    \caption{\small Re-Rank trajectories for SRLM on SafeRLHF. Rows show oracle-aware
    and oracle-free construction; dashed lines show clean SRO. Blue circles
    denote AQ and red squares SRF on a shared scale.}
    \label{fig:oracle-free-main-srlm-rerank}
\end{figure*}

\begin{figure*}[t]
    \centering
    \includegraphics[width=0.97\textwidth]
        {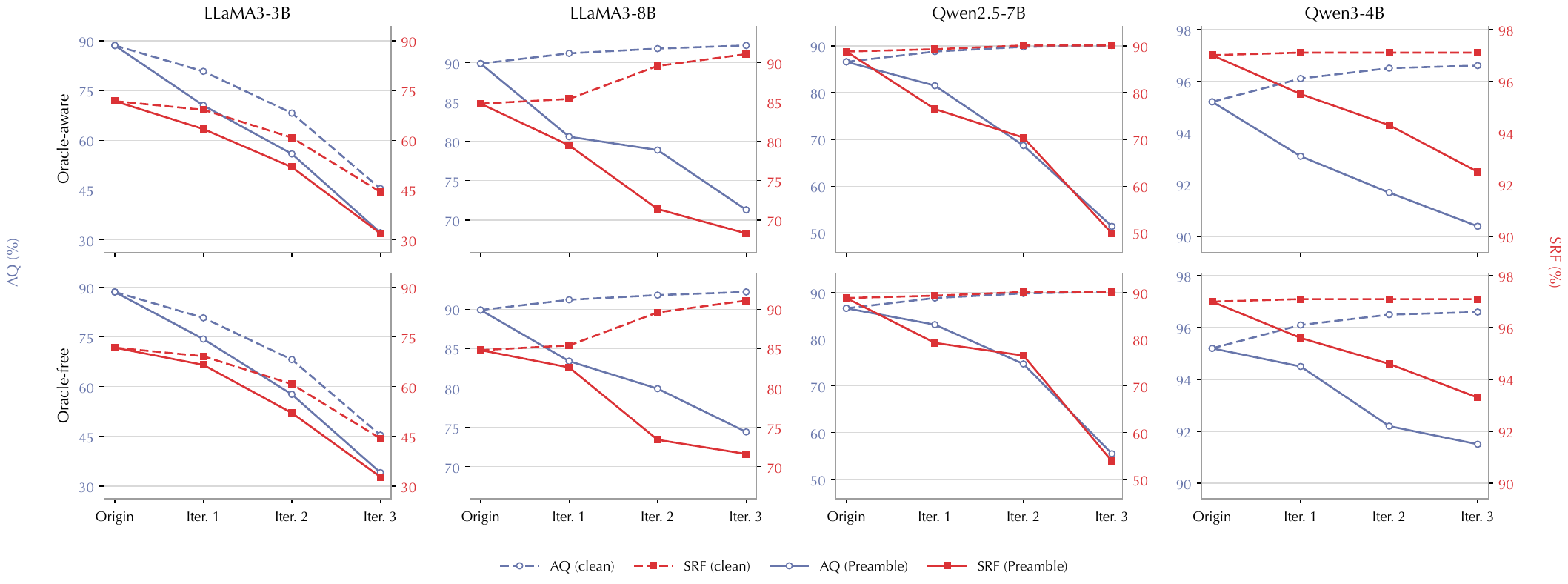}
    \caption{\small Preamble trajectories for SRLM on SafeRLHF.}
    \label{fig:oracle-free-main-srlm-preamble}
\end{figure*}

\begin{figure*}[t]
    \centering
    \includegraphics[width=0.97\textwidth]
        {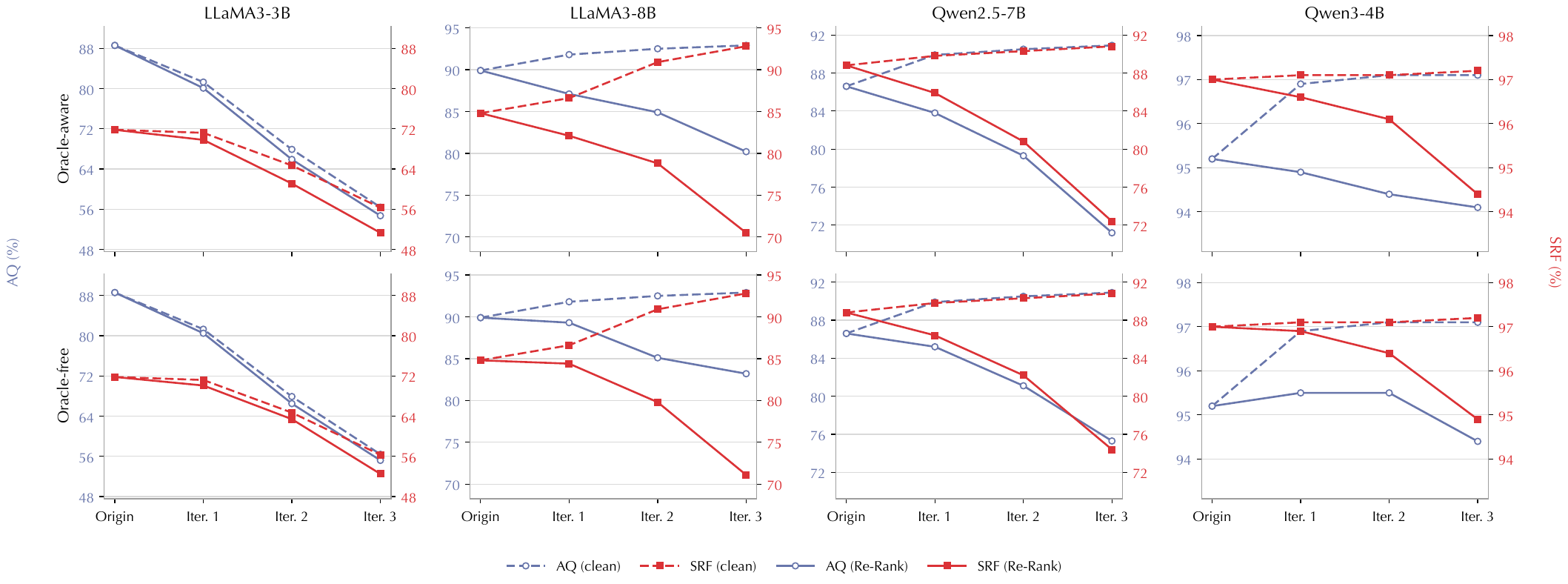}
    \caption{\small Re-Ranking trajectories for CREAM on SafeRLHF.}
    \label{fig:oracle-free-main-cream-rerank}
\end{figure*}

\begin{figure*}[t]
    \centering
    \includegraphics[width=0.97\textwidth]
        {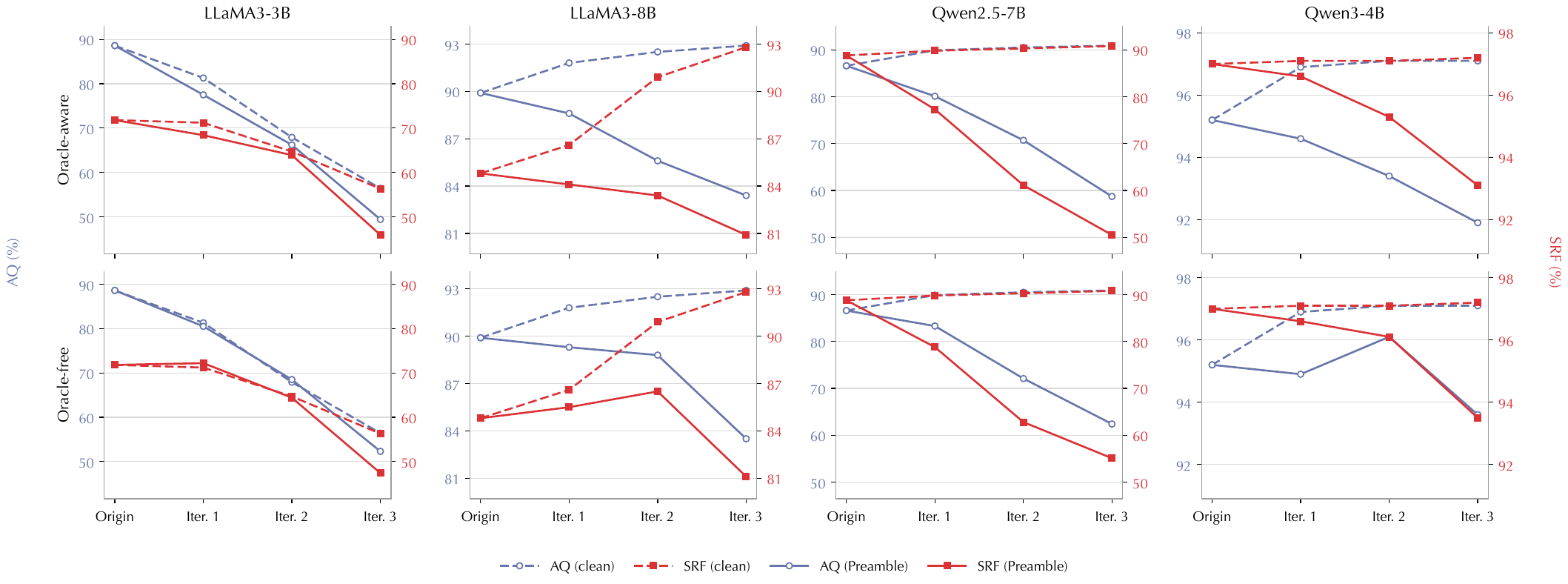}
    \caption{\small Preamble trajectories for CREAM on SafeRLHF.}
    \label{fig:oracle-free-main-cream-preamble}
\end{figure*}

Figures~\ref{fig:oracle-free-main-srlm-rerank}--\ref{fig:oracle-free-main-cream-preamble}
show the SafeRLHF trajectories. Oracle-free attacks are consistently weaker:
at every matched iteration, their AQ and SRF are no lower than those of the
corresponding oracle-aware attacks. The oracle is used only for post-hoc
evaluation of the oracle-free runs.

\subsection{Cross-Model Transfer With and Without an Oracle}
\label{sec:oracle-free-transfer}

We construct each attack once on a surrogate and reuse it for three SRLM
iterations on an unseen Llama-3.1-8B target, without querying the target during
construction. We compare a same-family surrogate (Llama-3.2-3B), a cross-family
surrogate (Qwen2.5-7B), and a target-matched reference. All runs use SafeRLHF
and the same settings; each Origin point is the target baseline.

\begin{figure*}[t]
    \centering
    \includegraphics[width=0.97\textwidth]
        {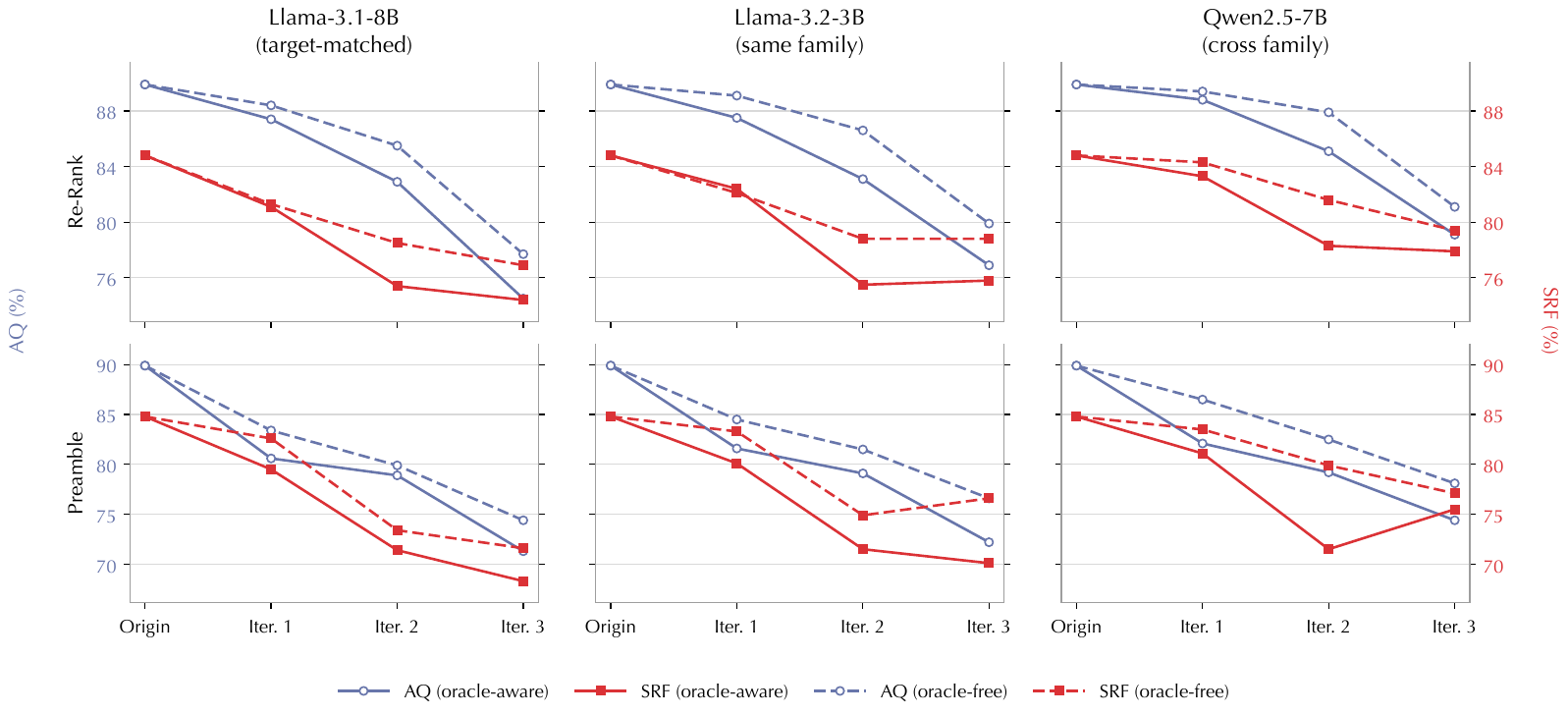}
    \caption{\small Cross-model transfer on SafeRLHF under SRLM. Rows show attacks,
    columns show surrogates, and solid/dashed lines denote oracle-aware/free
    construction. The Llama-3.1-8B column is a target-matched reference.}
    \label{fig:oracle-free-transfer}
\end{figure*}

As shown in Figure~\ref{fig:oracle-free-transfer}, oracle-aware attacks
transfer slightly more strongly: oracle-free AQ and SRF
are higher at every point except one $0.3$-point SRF difference. Same-family
transfer is also stronger than cross-family transfer for both attacks and
access settings at iteration 3. However, the final AQ gap is only $1.2$--$2.2$
points, and the comparison uses only one target and two surrogates.

\subsection{Sensitivity to Oracle Choice}
\label{sec:oracle-choice}

\begin{figure*}[t]
    \centering
    \includegraphics[width=0.82\textwidth]
        {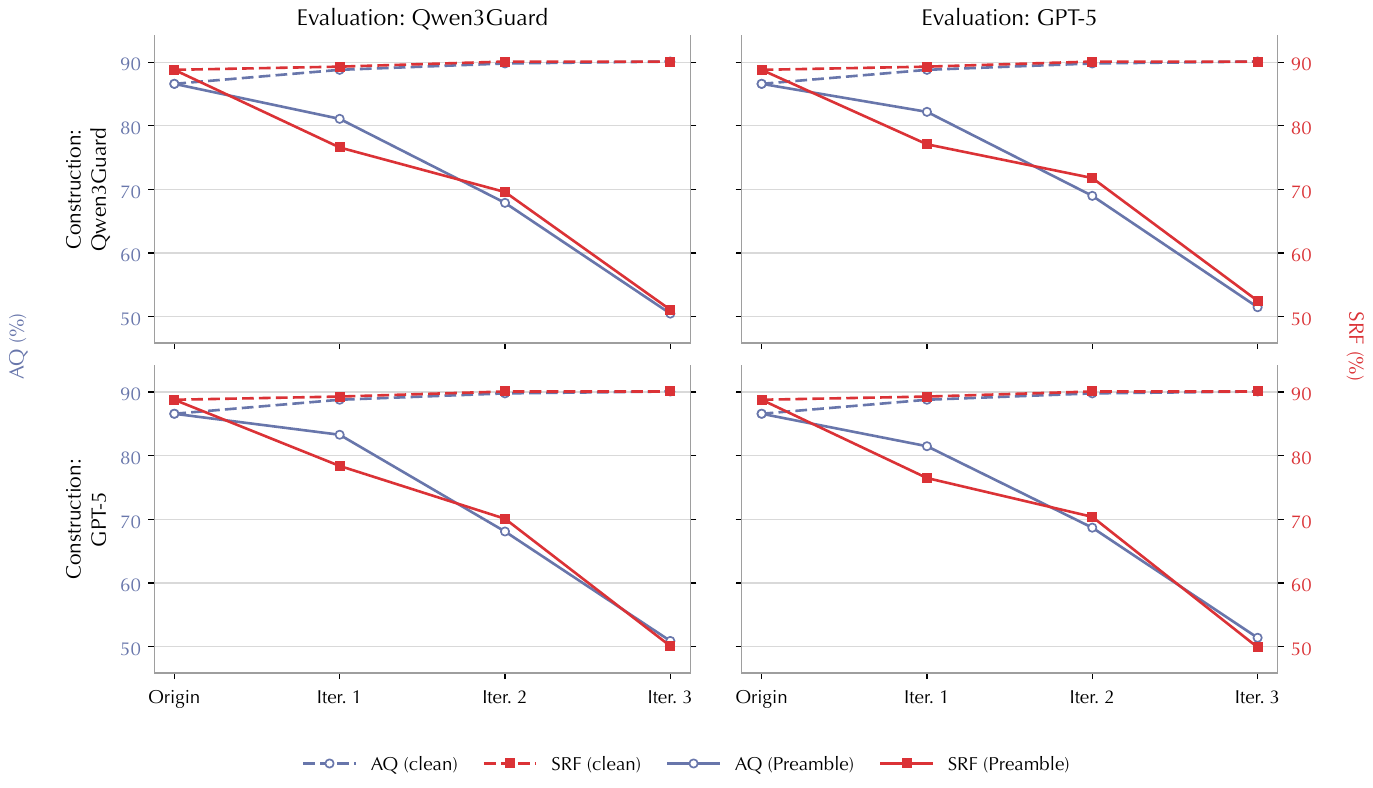}
    \caption{\small Preamble sensitivity to construction and evaluation oracles.
    Rows vary construction and columns evaluation. The dashed clean reference
    is the GPT-5-scored run. Qwen3Guard abbreviates
    Qwen3Guard-Gen-0.6B.}
    \label{fig:oracle-choice}
\end{figure*}

We independently vary the construction and evaluation oracles between GPT-5
and Qwen3Guard-Gen-0.6B. Figure~\ref{fig:oracle-choice} reports all four
pairings for Preamble on Qwen2.5-7B with SafeRLHF and SRLM.

All four pairings collapse similarly: final AQ is $50.5\%$--$51.5\%$ and SRF
is $49.9\%$--$52.5\%$, compared with $90.1/90.1$ for clean SRO. Changing the
construction oracle changes final AQ/SRF by at most $0.4/2.6$ points; changing
the evaluation oracle changes them by at most $1.0/1.4$ points. The
off-diagonal pairings therefore show that construction and evaluation need not
use the same oracle. This result is limited to one attack, model, and dataset.

\section{Frozen-Judge Controls}
\label{sec:frozen-judge-ablation}

A frozen self-judge isolates evaluator co-evolution from ordinary preference
optimization. The policy remains trainable and generates fresh on-policy
responses, but a fixed evaluator supplies the preference labels.

\subsection{Attack Amplification}

\paragraph{Setup.}
We compare standard SRO with a frozen-judge control on SafeRLHF using
Llama-3.1-8B and SRLM. Standard SRO uses the current model as the judge,
whereas the control retains the initial judge:
\begin{equation}
    r_t^{\mathrm{SRO}}=r_{\theta_t},
    \qquad
    r_t^{\mathrm{frozen}}=r_{\theta_0}.
\end{equation}
Both variants use the same initialization, attack, and optimization protocol.

\begin{figure*}[t]
    \centering
    \includegraphics[width=0.82\textwidth]
        {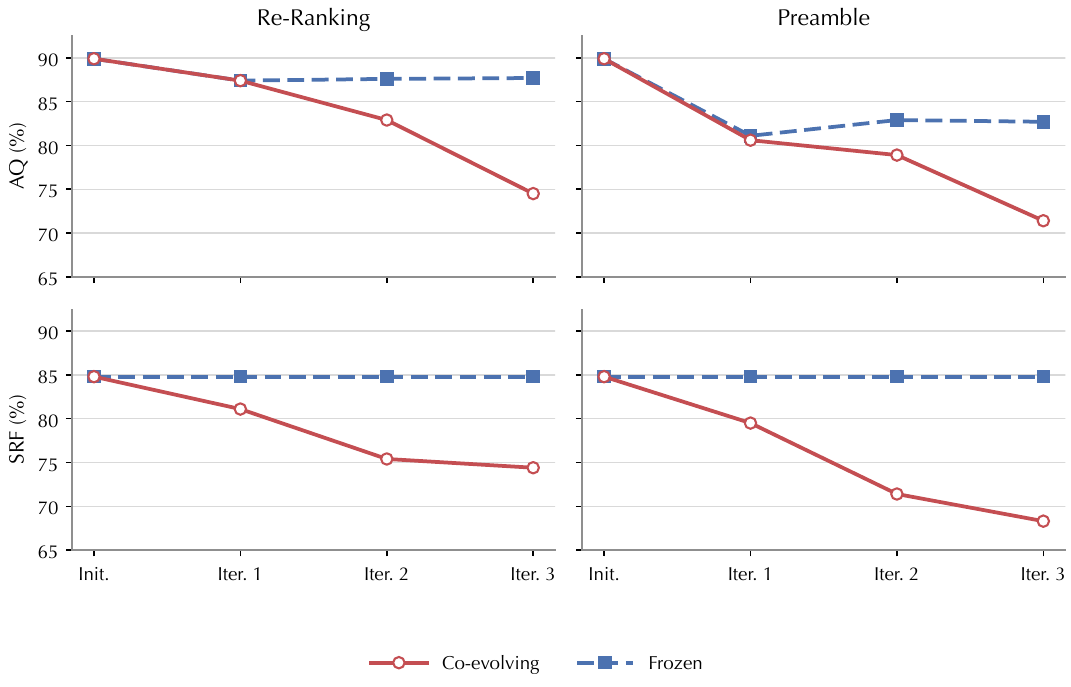}
    \caption{\small Attack trajectories with SRO and frozen self-judges.
    Both attacks cause a first-round AQ drop; the SRO trajectories
    continue to decline, while the frozen trajectories stabilize.}
    \label{fig:frozen-judge-attack-trajectories}
\end{figure*}

\paragraph{Results.}
Figure~\ref{fig:frozen-judge-attack-trajectories} shows an initial AQ drop
with either judge: Re-Ranking loses $2.5$ points in both conditions, while
Preamble loses $9.3$ points with the co-evolving judge and $8.8$ with the
frozen judge. From Iteration~1 to~3, AQ changes by $-12.9$ versus $+0.3$
points under co-evolving versus frozen Re-Ranking, and by $-9.3$ versus
$+1.6$ points under Preamble. Thus, the attacks are not specific to SRO, but
their progressive damage is amplified when policy updates also modify the
judge that supervises the next round.

\subsection{Controlled-SRF Amplification}
\label{sec:frozen-judge-srf}

The controlled-SRF experiment in Section~4.1.1 shows that SRO changes from
improvement to degradation as self-generated supervision becomes less
reliable. However, corrupted preference labels can also harm ordinary
preference optimization with a fixed evaluator. We therefore compare standard
SRO with a frozen self-judge to separate the direct effect of low-fidelity
preferences from their subsequent amplification through self-reward feedback.

\paragraph{Matched first update.}
For each perturbation level $\epsilon$, both branches start from the same
initial model and use the same first-round prompts, sampled responses,
perturbed reward labels, preference pairs, and optimization settings. Positive
$\epsilon$ corrupts a controlled fraction of initially correct labels, while
negative $\epsilon$ corrects a fraction of initially incorrect labels. We
report the realized fidelity of this perturbed supervision as the initial SRF.

The two branches also use paired random seeds, so they produce the same
post-update checkpoint up to optimization stochasticity and share
$\mathrm{AQ}_1$. Synthetic label perturbation is applied only in this first
round. Thus, the first update measures the immediate effect of low-fidelity
preference optimization; later divergence measures how this initial error
propagates through the training loop.
\paragraph{Dynamic versus frozen self-judges.}
After the matched first update, standard SRO uses the current model as its
evaluator, while the frozen branch retains a separate copy of the initial
evaluator:
\begin{equation}
    r_t^{\mathrm{SRO}}=r_{\theta_t^{\mathrm{SRO}}},
    \qquad
    r_t^{\mathrm{frozen}}=r_{\theta_0},
    \qquad t\geq 1.
\end{equation}
The frozen policy remains fully trainable and generates fresh on-policy
responses at every iteration; only the evaluator parameters are fixed. The
prompt sequence, number of responses, decoding configuration,
preference-optimization objective, learning rate, and number of outer
iterations are otherwise matched.

\paragraph{Evaluation.}
Because both branches share the first corrupted update, we quantify subsequent
propagation using
\begin{equation}
    \Delta_{\mathrm{post}}
    = \mathrm{AQ}_3-\mathrm{AQ}_1.
    \label{eq:post-first-change}
\end{equation}
Positive values indicate improvement after the common update; negative values
indicate additional degradation. A substantially more negative
$\Delta_{\mathrm{post}}$ under SRO than under the frozen judge is evidence that
evaluator co-evolution amplifies the initial preference corruption.

\begin{figure*}[t]
    \centering
    \includegraphics[width=0.7\textwidth]
        {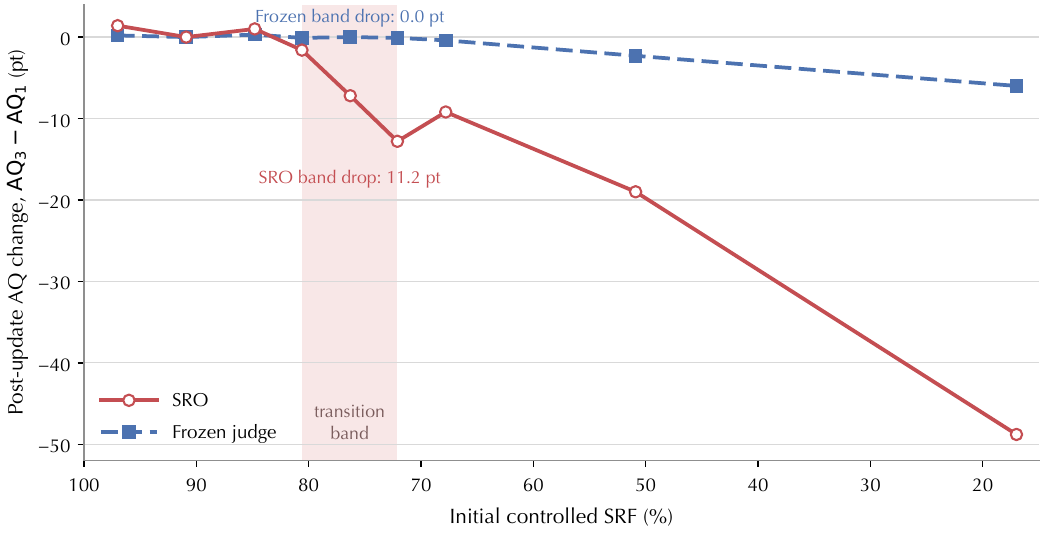}
    \caption{\small Post-first-update AQ change versus controlled initial SRF.
    Fidelity decreases from left to right, and
    $\mathrm{AQ}_3-\mathrm{AQ}_1$ isolates change after the matched first
    update. The shaded region spans $80.6\%$--$72.1\%$ initial SRF. Across
    this interval, the SRO change falls by $11.2$ points, whereas the
    frozen-judge change remains at $-0.1$ points.}
    \label{fig:frozen-judge-aq-change}
\end{figure*}

\begin{figure*}[t]
    \centering
    \includegraphics[width=0.98\textwidth]
        {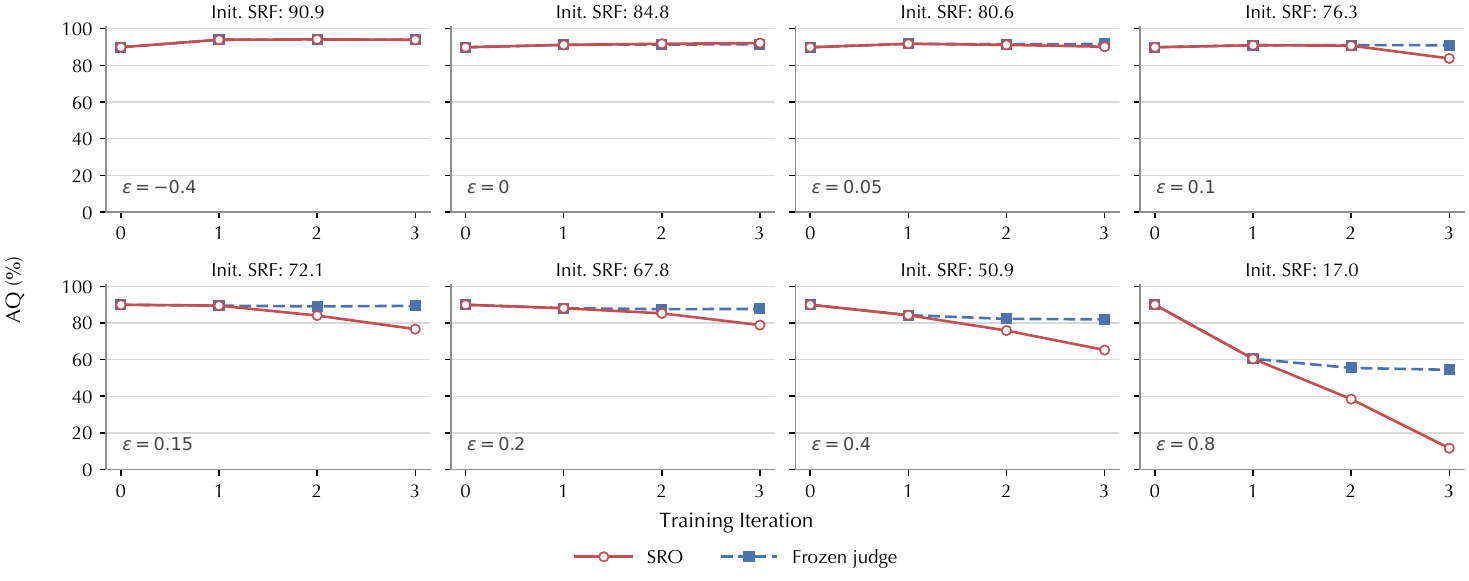}
    \caption{\small AQ trajectories across controlled perturbation levels. The performance of SRO and frozen self-judge in subsequent iterations diverge as initial SRF decreases.}
    \label{fig:frozen-judge-trajectories}
\end{figure*}

\paragraph{Complementary views.}
Figures~\ref{fig:frozen-judge-aq-change} and
\ref{fig:frozen-judge-trajectories} summarize the same comparison. The first
aggregates post-first-update change over all controlled fidelity levels; the
second shows the complete Iteration~0--3 AQ trajectories.

\paragraph{Results and interpretation.}
At high initial SRF, both branches preserve or improve AQ after the shared
first update. Near the transition region, the trajectories separate:
$\Delta_{\mathrm{post}}$ is $-1.6$, $-7.2$, and $-12.8$ points under SRO at
$80.6\%$, $76.3\%$, and $72.1\%$ initial SRF, respectively, but remains
between $-0.1$ and $0.0$ points with the frozen judge. At lower fidelity, both
methods can degrade, but the SRO decline is substantially steeper. For
example, at $67.8\%$ initial SRF, SRO loses $9.2$ points after the common
update, compared with $0.4$ points for the frozen baseline; at $17.0\%$, the
corresponding losses are $48.8$ and $6.0$ points.

The shared change from Iteration~0 to~1 captures the direct cost of optimizing
against perturbed preferences. The growing gap after Iteration~1 instead
isolates the effect of allowing the evaluator to co-evolve with the policy.
Together, the two figures show that low-fidelity supervision can harm either
branch initially, but the characteristic progressive collapse is much
stronger when each policy update also changes the judge that generates the
next round of supervision.